\documentclass{article}

\usepackage[preprint]{neurips_2021}

\usepackage[utf8]{inputenc} 
\usepackage[T1]{fontenc}    
\usepackage{hyperref}       
\usepackage{url}            
\usepackage{booktabs}       
\usepackage{amsfonts}       
\usepackage{nicefrac}       
\usepackage{microtype}      
\usepackage{xcolor}         
\usepackage{wrapfig}
\usepackage{adjustbox}
\usepackage{pifont}
\usepackage{multirow}
\usepackage{array}
\usepackage{amsmath}
\usepackage{amsthm}
\usepackage{lipsum}
\usepackage{cases}
\usepackage{mathtools}
\usepackage{tikz}
\usepackage{algorithm}
\usepackage{algorithmic}
\usepackage{forloop}
\usepackage[font=small,skip=5pt]{caption}
\usepackage{moreenum}
\usepackage{cleveref}

\newcommand{\rv}[1]{\mathbf{#1}}
\newcommand{\data}{\mathcal{D}}
\newcommand{\dkl}[2]{\mathrm{D}_{\mathtt{KL}}({#1}||{#2})}
\newcommand{\rvx}{\mathbf{x}}
\newcommand{\rvz}{\mathbf{z}}
\newcommand{\rvy}{\mathbf{y}}
\newcommand{\rvu}{\mathbf{u}}
\newcommand{\rva}{\mathbf{a}}
\newcommand{\loge}{\mathrm{log}}

\newcommand{\info}[2]{\mathrm{I}(\mathbf{#1},\mathbf{#2})}
\newcommand{\infos}[2]{\mathrm{I^*}(\mathbf{#1},\mathbf{#2})}
\newcommand{\hinfo}[2]{\hat{\mathrm{I}}(\mathbf{#1},\mathbf{#2})}
\newcommand{\infoq}[2]{\mathrm{I}_\mathrm{q}(\mathbf{#1},\mathbf{#2})}
\newcommand{\infop}[2]{\mathrm{I}_\mathrm{p}(\mathbf{#1},\mathbf{#2})}
\newcommand{\hinfoq}[2]{\hat{\mathrm{I}}_q(\mathbf{#1},\mathbf{#2})}

\newcommand{\entropy}[1]{\mathrm{H}(\mathbf{#1})}
\newcommand{\entropyp}[1]{\mathrm{H}_\mathrm{p}(\mathbf{#1})}
\newcommand{\entropyq}[1]{\mathrm{H}_\mathrm{q}(\mathbf{#1})}

\newcommand{\expect}[2]{\mathrm{E}_{#1} \big[ #2 \big]}
\newcommand{\expectbig}[2]{\mathrm{E}_{#1} \bigg[ #2 \bigg]}

\newcommand{\pthe}{p_\theta}
\newcommand{\qphi}{q_\phi}
\newcommand{\qdata}{q_\data}
\newcommand{\qxz}{q_{\phi}}
\newcommand{\qvf}{q_\varphi}
\newcommand{\ppsi}{p_\psi}
\newcommand{\qtau}{q_\tau}
\newcommand{\pkap}{p_\kappa}

\newcommand{\elbo}{\mathcal{E}}
\newcommand{\elbou}{\mathcal{E}_\mathtt{U}}
\newcommand{\elbos}{\mathcal{E}_\mathtt{S}}
\newcommand{\elbosemafo}{\mathcal{E}_{\mathtt{SemafoVAE}}}

\newcommand{\loss}{\mathcal{L}}

\newtheorem{theorem}{Theorem}

\newtheorem{lemma}[theorem]{Lemma}

\theoremstyle{definition}
\newtheorem{definition}{Definition}[section]

\theoremstyle{remark}

\newtheorem{proposition}{Proposition}

\title{The Transitive Information Theory and its Application to Deep Generative Models}

\author{
  Trung Ngo \\
  School of Computing \\
  University of Eastern Finland \\
  Joensuu, Finland \\
  \texttt{trung@uef.fi} \\
  \And
  Najwa Laabid \\
  School of Medicine \\
  University of Eastern Finland \\
  Kuopio, Finland \\
  \texttt{najwa.laabid@uef.fi} \\
  \And
  Ville Hautam\"{a}ki \\
  School of Computing\\
  University of Eastern Finland\\
  Joensuu, Finland \\
  \texttt{villeh@cs.uef.fi} \\
  \And
  Merja Hein\"{a}niemi \\
  School of Medicine\\
  University of Eastern Finland\\
  Kuopio, Finland \\
  \texttt{merja.heinaniemi@uef.fi} \\
}

\begin{document}

\maketitle

\begin{abstract}
  Paradoxically, a Variational Autoencoder (VAE) could be pushed in two opposite directions, utilizing powerful decoder model for generating realistic images but collapsing the learned representation, or increasing regularization coefficient for disentangling representation but ultimately generating blurry examples. Existing methods narrow the issues to the rate-distortion trade-off between compression and reconstruction. We argue that a good reconstruction model does learn high capacity latents that encode more details, however, its use is hindered by two major issues: the prior is random noise which is completely detached from the posterior and allow no controllability in the generation; mean-field variational inference doesn't enforce hierarchy structure which makes the task of recombining those units into plausible novel output infeasible. As a result, we develop a system that learns a hierarchy of disentangled representation together with a mechanism for recombining the learned representation for generalization. This is achieved by introducing a minimal amount of inductive bias to learn controllable prior for the VAE. The idea is supported by here developed transitive information theory, that is, the mutual information between two target variables could alternately be maximized through the mutual information to the third variable, thus bypassing the rate-distortion bottleneck in VAE design. In particular, we show that our model, named SemafoVAE (inspired by the similar concept in computer science), could generate high-quality examples in a controllable manner, perform smooth traversals of the disentangled factors and intervention at a different level of representation hierarchy.
\end{abstract}

\section{Introduction}
\label{sec:intro}


Earlier effort in generative model was solely relied on statistical model defined by human experts, inference for such model is tractable by narrow a set of strict assumption regarding the data generation process \cite{bishop06_Patternrecognitionmachine}. Conversely, modern methods leverage recent advance in computing to approximate the generation process using powerful nonlinear model and big data. The two prominent families of these methods are: implicit generative model such as generative adversarial network (GAN) \cite{goodfellow14_Generativeadversarialnets} and explicit generative model which includes the variational autoencoder (VAE) \cite{kingma14_AutoEncodingVariationalBayes}. While the first approach have demonstrated its merits in generating realistic high-quality image \cite{karras2018progressive}, the second one is often referred as a representation learning algorithm that capturing independent \textit{factor of variations} (FOVs) also known as disentanglement representation \cite{higgins17_betaVAELearningBasic,locatello19_ChallengingCommonAssumptions}.

The benefit of learning independent generative factors are discussed in \cite{bengio14_RepresentationLearningReview} and \cite{scholkopf21_causalrepresentationlearning}, these include: boosting the performance of downstream task, improving the robustness of generative model under distribution shift and discovering the causal variables. Thus, we could reasonably assume that a model that learn relevant factors for generating data would have better understanding of the data manifold by itself, subsequently, enabling it to generate better images. However, this isn't the case for the known families of generative methods. First, GAN doesn't explicitly learn a meaningful representation, the whole generation process is distilled into the deep generator network which have been known to suffer from mode collapse issue \cite{goodfellow17_NIPS2016Tutorial}. In contrast, VAE has established to be a reliable performer under various disentanglement representation benchmarks \cite{locatello19_ChallengingCommonAssumptions,qiao19_DisentanglementChallengeRegularization}, and its ability to learn a tractable latent distribution enables the representation to be generalized beyond the reconstruction task. This capability doesn't come without a drawback, pushing the compression rate in VAE is equivalent to forcing the high-distorted outputs \cite{alemi17_FixingBrokenELBO}, as a result, the generated image is blurry and lacks detail \cite{burgess18_UnderstandingdisentanglingvVAE}.

In practice, VAE is capable of generating high-fidelity images by carefully redesign its architecture \cite{maaloe19_BIVAverydeep,vahdat20_NVAEDeepHierarchical,child21_VerydeepVAEs}. These designs significantly increase the depth both VAE's encoder and decoder, and allows the accommodation of the hierarchical latent variables. It is unclear how the complication of design would affect the ability to learn independent meaningful factors of VAE, and these models haven't been evaluated against the \textit{state-of-the-art} (SoTA) disentangling methods \cite{locatello19_ChallengingCommonAssumptions}. Preliminary studies in \cite{havtorn21_HierarchicalVAEsknow} indicates that hierarchical VAE does learn hierarchical representation by adding layers of fine details to the mode of learned distribution, however, this raises more important question about how to navigate through a large number of possible latent units combinations to sample the \textit{attributes of interest}. To revisit the initial claim, we reason that generative model doesn't need to generate all the possible images but only the images with particular attributes in a controllable manner for real life setting. Additionally, our lack of understanding of such high-resolution representation is apparently the missing of a learnable controlling mechanism for generation \cite{montero21_roledisentanglementgeneralisation}, i.e. a compositional mechanism that recombining disentangled and hierarchical representation in a meaningful way.

The method presented in this paper addressing three major issues with the conventional VAE framework: 1) learning the hierarchy of factors that are disentangled; 2) learning the compositional mechanism to control the learned representation and 3) all these developments are achieved while improving the expressiveness of VAE generator. In summary, our contributions are following:
\begin{enumerate}
  \item We provide theoretical and empirical justification for the limitation of VAE framework (Section~\ref{sec:background}).
  \item We develop the transitive information theory explaining how information is transferred among variables. Based on the proposed principles, we implement semi-supervised ``SemafoVAE'' that encapsulate variables' hierarchy in its prior and allow explicit control of the generation. (Section~\ref{sec:method}).
  \item The algorithm is benchmarked against the SoTAs in terms of test log-likelihood, generation quality and disentanglement metrics (Section~\ref{sec:experiments}).
\end{enumerate}

\section{Background: VAE and its limitations}
\label{sec:background}

In this section, we review prior work and discuss the VAE's limitations as a method for generative modeling and representation learning.

\subsection{Variational autoencoder}

Variational autoencoder~\cite{kingma14_AutoEncodingVariationalBayes} introduces the latent variables $\rvz$ that enables learning richer representation of the observation $\rvx$. In the latent variable framework, we obtain marginal $\pthe(\rvx) = \int_{\rvz} \pthe(\rvx|\rvz)p(\rvz) d\rvz$, however, the marginalization of the likelihood is intractable. Variational method approximates the posterior distribution $\pthe(\rvz|\rvx)$ with a tractable distribution $\qphi(\rvz)$, and treats the issue of closing the approximation gap as an optimization problem w.r.t the parameters $\phi$. However, the latent variables are optimized per-data point which is another obstacle for scaling the algorithm. Instead, amortized inference learns the mapping  $\qphi(\rvz|\rvx)$, and added to the scalability is stochastic optimization for minimizing the posterior divergence $\dkl{\qphi(\rvz|\rvx)}{\pthe(\rvz|\rvx)}$. The log-likelihood of our data is decomposed into two terms as in \cite{kingma14_AutoEncodingVariationalBayes},
\begin{equation}
  \begin{aligned}
    \log \pthe(\rvx) \ge
    \underbrace{
      \overbrace{\expect{\qphi(\rvz|\rvx)}{\loge \pthe(\rvx|\rvz)}}^{reconstruction} -
      \overbrace{\dkl{\qphi(\rvz|\rvx)}{p(\rvz)}}^{regularisation}
    }_{=\elbo(\rvx,\rvz;\phi,\theta)},
  \end{aligned}
  \label{eq:elbo}
\end{equation}
where $\elbo(\rvx,\rvz;\phi,\theta)$ defines the \textit{evidence lower bound} (ELBO), which is maximized for each data point w.r.t the parameters using Monte Carlo estimate and stochastic gradient descent (SGD). The VAE~\cite{kingma14_AutoEncodingVariationalBayes} uses \textit{reparameterization trick} to jointly optimize $\qphi(\rvz|\rvx)$ and $\pthe(\rvx|\rvz)$, where $\phi$ and $\theta$ are parameters of two deep neural networks, and the combination of these techniques enables variational inference (VI) to be both flexible and scalable.

Throughout this study, we denote $\rvx$ as random variables (RVs) represent the observations which are i.i.d samples from the dataset $\data$ with empirical data distribution $q_\data(\rvx) = \frac{1}{N} \sum_i^N \delta(x_i)$. Then, $\rvz$ as the latent RVs, and $\rvy$ are the ground truth factors often understood as the true low-dimensional manifold embedding of $\rvx$ \cite{dai19_DiagnosingEnhancingVAE}. For the clarity of notation, our derivations in the next sections will omit the parameters $\phi$ and $\theta$.

\subsection{Rethinking the ELBO objective}
\label{subsec:limit_elbo}

\begin{wrapfigure}{r}{0.4\textwidth}
  \vspace{-10pt}
  \begin{small}
    \begin{center}
      \includegraphics[width=\linewidth]{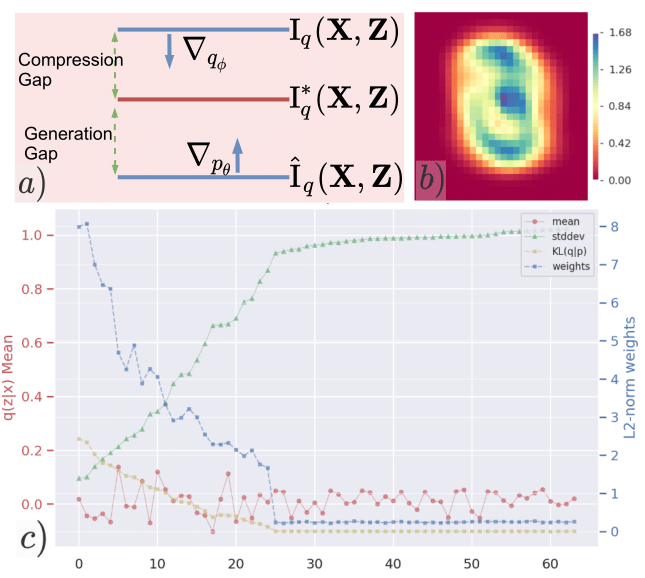}
    \end{center}
    \caption{$a)$ illustration of the competing objectives between the encoder $\qphi$ and decoder $\pthe$ of the VAE, where $\nabla$ indicates the corresponding gradient direction, and $\mathrm{I}^*_q(\rv{x};\rv{z})$ is the optimal balance between compression and generation quality. $b)$ heatmap of reconstruction negative log-likelihood on MNIST test set, \textit{red color indicate smaller values}. $c)$ shows the latent units' statistics of VAE trained on MNIST, \textit{weights} are the L2-norm of decoder weights placed for each individual latent units (there are 64 units in total, and 25 units are activated).}
    \label{fig:adversarial}
  \end{small}
  \vspace{-10pt}
\end{wrapfigure}

\begin{lemma}
  For any encoder model $q(\rvz|\rvx)$ and decoder model $p(\rvx|\rvz)$, optimizing ELBO is equal to the minimization of
  \begin{equation}
    \begin{aligned}
      \loss(\data,\phi,\theta)
    & = \underbrace{
          \expectbig{
            q(\rvx,\rvz) }{
            \loge \frac{q(\rvx,\rvz)}{q(\rvx)q(\rvz)} }
        }_{=\infoq{x}{z}\; \mathrm{(a)}} -
        \underbrace{
          \expectbig{
            q(\rvx,\rvz) }{
            \loge \frac{p(\rvx,\rvz)}{q(\rvz)q(\rvx)} }
        }_{=\hinfoq{x}{z}\; \mathrm{(b)}} -
        \underbrace{
          \expectbig{ q(\rvx) }{ \loge q(\rvx) }
        }_{\text{fixed}}
    \end{aligned}
    \label{eq:elbo_adversarial}
  \end{equation}
  which jointly:
  i) pushes the encoder to compress the latent codes by disregarding observational information; and
  ii) recovers missing information in the latent codes using the generator. (Proof is in the Appendix~\ref{app:subsec:elbo_adversarial})
  \label{lem:elbo_adversarial}
\end{lemma}

First,~\cref{lem:elbo_adversarial} indicates maximizing ELBO is the equivalent of minimizing the posterior mutual information $\infoq{x}{z}$ (term $a)$ in~\eqref{eq:elbo_adversarial}. This fact was first mentioned in \cite{hoffman16_ELBOsurgeryanother}, and it is indeed the desire property of ELBO that facilitate generalization as interpreted by \cite{alemi17_FixingBrokenELBO,alemi19_DeepVariationalInformation}. According to the information bottleneck principle \cite{tishby99_informationbottleneckmethoda}, the ELBO is $\mathrm{max} (\info{x}{z} - \beta \info{z}{i})$ where $\mathbf{i}$ is the index to the individual example. However, this interpretation leaves much to be answered since the VAE could solely focus on learning the mean of $\rvx$ if required to ignore all the individual details. Paradoxically, study in \cite{bozkurt_RateRegularizationGeneralizationVAEs} concluded that the best generalized VAE was achieved by severely weakening the KL-regularization term in \eqref{eq:elbo}, while other studies in \cite{higgins17_betaVAELearningBasic,burgess18_UnderstandingdisentanglingvVAE,higgins18_SCANLearninghierarchical,montero21_roledisentanglementgeneralisation} proposes an opposite approach that increasing regularization of the latents would encourage VAE to learn more generalized representation.

We prove that term $b)$ in \eqref{eq:elbo_adversarial} is actually the lower bound of $a)$, i.e. $\hinfoq{x}{z} \le \infoq{x}{z}$. Since the last quantity is constant for a given dataset, the ELBO \textit{game} focuses on the interaction between $a)$ - the encoder and $b)$ - the decoder. Figure~\ref{fig:adversarial}-a illustrates both players have opposed objective to realize their maximum capacity at the optimal point $\infos{x}{z}$. The description resembles an \textit{adversarial game}, in practice, encoder and decoder coordinate together reaching an equilibrium point. If $\infoq{x}{z}$ is small (i.e. over-regularized VAE), the decoder receives no update because its objective is easily reached, as a result, it is saturated to the maximum entropy of the output distribution which results blurry images. An equivalent observation applied for VAE with powerful decoder or compromised regularization, $\hinfoq{x}{z}$ would reach its maximum and move the optimal point $\infos{x}{z}$ up, which narrows down the compression gap and stops encoder from obtaining meaningful codes. To this end, we argue that the original ELBO objective is capable of achieving optimal equilibrium for both representation learning and generative modeling, however, this is often hampered by flaw in optimization algorithm. This is corroborated by experiments in Section~\ref{sec:experiments} and also additional experiments in the Appendix.

\subsection{Limitation of the maximum likelihood estimation}
\label{subsec:limit_mle}

The limitation of MLE has been studied for decades \citep{bishop06_Patternrecognitionmachine}, and Figure~\ref{fig:adversarial}-b shows that deep learning is no exception when MLE is used as an objective. Most of the learning of MLE involves pushing the marginal density area to zero, and the approximated density is severely limited by biases within training data. Since our analysis in the previous Section indicates a good decoder needed for good representation, it is understandable why encoder is suboptimal at the beginning of VAE training~\cite{sonderby16_Laddervariationalautoencoders,kingma16_Improvedvariationalinference,he19_Lagginginferencenetworks}. According to this hypothesis, VAE with MLE objective is impossible to achieve extrapolation, since any non-zero pixels in the \textit{red zone} have zero likelihood. A similar observation is also empirically validated in \cite{montero21_roledisentanglementgeneralisation}. We suspect that the \textit{dead pixels} issue is closely associated to posterior collapse in VAE \cite{kingma16_Improvedvariationalinference,lucas19_UnderstandingPosteriorCollapsea,dai19_UsualSuspectsReassessing} as the collapsed latent units could be used to specify the location of invariant pixels for the whole training set. As a result, we add ``free-pixels'' to the reconstruction term to prevent the over penalization of empty pixels, i.e. $\elbo(\rvx,\rvz) = (\expect{q(\rvz|\rvx)}{p(\rvx|\rvz)} + \mathrm{R}) - \dkl{q(\rvz|\rvx)}{p(\rvz)}$ where $\mathrm{R}$ is a chosen coefficient, the idea is similar to the ``free-bits'' approach in \cite{kingma16_Improvedvariationalinference}. However, the extra constraint only delays onset of likelihood saturation and powerful deep network is perfectly capable of adjusting its threshold values to compensate the fixed density (extra results in the Appendix).

\subsection{Limitation of the deep autoencoder architecture}
\label{subsec:limit_ae}

We find that the autoencoder architecture doesn't allow VAE to utilize all of its latent units since there is an upper bound for the information that passes the bottleneck.

\begin{definition}[$\alpha$-active VAE]
  A VAE with $\alpha$ number of latent units that don't collapse to the prior given a sufficient number of latent dimensions, so that there is at least one latent unit $j$ that is collapsed, i.e. $q(z_j|\rvx) = p(z_j)$.
\end{definition}

\begin{proposition}
  Any VAE trained on the same dataset, with the similar capacity for the encoder and decoder, and the same choice of distributions for the posterior, prior and likelihood belong to the same family of $\alpha$-active VAEs, regardless the number latent units in the bottleneck or the amount of training data.
  \label{prop:active_units}
\end{proposition}

\begin{proposition}
  A trained $\alpha$-active VAE encoder places an upper bound reconstruction quality for any decoder that is trained using its learned representation.
  \label{prop:decoder_bound}
\end{proposition}

Discussion and empirical proof are in the Appendix. In ~\cite{dai19_UsualSuspectsReassessing}, the authors argue that the posterior collapse is a direct consequence of local minima of deep autoencoder networks. However, the networks used in practice is far more complicated than the networks with soft-threshold activation, and the recent theory in \cite{nakkiran19_DeepDoubleDescent} suggests deep networks are beneficial in avoiding overfitting to local minima. Figure~\ref{fig:adversarial}-c shows that VAE with capable encoder and decoder has exactly \textit{25} activated latent units for MNIST, this number remains consistent for any number of latent dimensions that is greater than \textit{25}. If a smaller number of dimensions is given, all units are activated and the reconstruction quality is reduced. This observation is repeated among multiple datasets, and the same phenomenon is observed on the same network trained with less amount of data, or with different choice for posterior, prior or reconstruction likelihood. The only way to change such balance is varing the capacity either encoder or decoder networks via the regularization weight ($\beta$-VAE~\cite{higgins17_betaVAELearningBasic}) or the network architectures.

If we assume that $\beta$-VAE learns a generalized disentangling factors and the weak decoder is a by-product of the process. Then a new capable decoder that is trained on the learned representation should be able to reconstruct a decent quality image. However, our experiments\footnote{in the Appendix} shows the fine-tuned decoder generated similar blurry output which suggests the encoder simply throws away information. Thus, the encoder will put an upper information bound to the decoder according to the \textit{data processing inequality}~\cite{cover06_Elementsinformationtheory}. This is the gist of Proposition~\ref{prop:decoder_bound} and consistent with our interpretation in Section~\ref{subsec:limit_elbo}. It is notable that not only the latent units are collapsed, the decoder also adapts its weights to zeros for the inactivated units Figure~\ref{fig:adversarial}-c. As a result, any attempt to revive the \textit{dead units} without restarting decoder will be fruitless.

\section{Method}
\label{sec:method}

So far we have only studied the ELBO objective and the interaction between $\rvx$ and $\rvz$, now we need to delve into $\rvy$ and its relation to latent $\rvz$. We formalize the relationship into a theorem in~\cref{subsec:info_trans}. Then, we propose a semi-supervised VAE algorithm, based on the developed theory.

\subsection{Information is transitive}
\label{subsec:info_trans}

\begin{theorem}[Transitive Information]
  For any set of three random variables $\rvx$, $\rvz$ and $\rvy$ so that $\entropy{y|x,z} \ge 0$:
  \begin{small}
    \begin{equation}
      \info{x}{z} \ge \info{x}{y} + \info{y}{z} -
      \mathrm{H}(\rvy)
      =\vcentcolon \hinfo{x}{z},
      \label{eq:trans_info}
    \end{equation}
  \end{small}
  where the equality is achieved when $\rvy \subset \rvz$ and
  $\info{x}{z} = \info{x}{y}$. (Proof relies on two properties of entropy $\entropy{x|z} \le \entropy{x,y|z}$ and $\entropy{x|y,z} \le \entropy{x|y}$, detailed derivation is in the Appendix~\ref{app:subsec:trans_info})
  \label{the:trans_info}
\end{theorem}

Theorem~\ref{the:trans_info} implies that maximizing MI can be transitive based on the choice of the random variable $y$, even though MI does not satisfy the \textit{triangle inequality}. The theorem is powerful in a sense that it is applied for any set of three random variables. As we could define $\rvy$ that is both computationally efficient and tractable, our algorithm could maximize the lower bound of the desired MI without having the access to the analytical solution. For instance, given $\rvx$ has 784 dimensions and $\rvz$ has 32 dimensions, we choose $\rvy$ with 10 dimensions, so that maximizing $\info{x}{y}$ and $\info{y}{z}$ involves iterating $784 \cdot 10 + 10 \cdot 32$ dimensions which is $3.07$ times faster than $\info{x}{z}$.

\begin{wrapfigure}{r}{0.5\textwidth}
\vspace{-10pt}
\begin{small}
  \begin{center}
    \includegraphics[width=\linewidth]{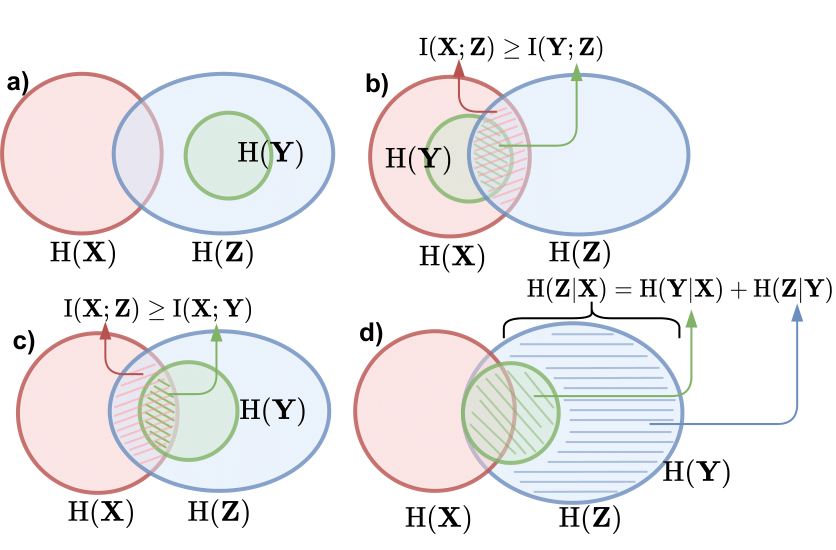}
  \end{center}
  \caption{Venn diagram of the transitive information theorem, illustrating interaction between $\rvx$ and $\rvz$ given different choices for $\rvy$: $a)$ $\rvy$ is random noise; $b)$ $\rvy$ is ground truth; $c)$ $\rvy$ is partially observed ground truth; and $d)$ achieved equality.}
  \label{fig:trans_info}
\end{small}
\vspace{-10pt}
\end{wrapfigure}

In practice, our concern is the interaction between $\rvy$ and $\rvz$. The first case is that $H(\rvy|\rvx)=0$ (Fig.~\ref{fig:trans_info}-b)), i.e. $\rvy \in \rvx$, then \eqref{eq:trans_info} takes simpler form $\info{x}{z} \ge \info{y}{z}$, hence, increasing $\info{y}{z}$ directly pushes the lower bound of our target MI. Because $\rvy$ and $\rvz$ are much lower dimensions than $\rvx$, the computational burden is significantly reduced. In the second case, our question is ``\textit{Would it be possible to learn latents that contain more information than the groundtruth factors?}'', in mathematical sense, it is $\info{x}{z} \ge \info{x}{y}$ (Fig.~\ref{fig:trans_info}-c)). This is proves to be possible when $H(\rv{y}|\rv{z}) = 0$ in which $\rvy \subset \rvz$ and the latents would absorb all the knowledge about $\rvy$, while being free to explore beyond the known manifold. The final case is when the solution for $\rvz$ is optimal, i.e. equality is achieved in~\eqref{eq:trans_info}. Figure~\ref{fig:trans_info}-d shows that $\rvz$ also contains all the information about $\rvy$, however, $\info{x}{z}$ is bounded to be equal to $\info{x}{y}$. Additionally, the role of $\rvy$ could be understood by the offset term $-H(\rvy)$, i.e. the more information we know about $\rvy$, the tighter the bound.

Theorem~\ref{the:trans_info} is an instrument for understanding the relationship between ground truth factors $\rvy$ and the learned latent $\rvz$. Specifically, it shows that there is an infinite number of solutions for learning $\rvz$ that achieved the same amount of MI with $\rvx$ as $\info{x}{y}$. Hence, a good representation might not need to be identical to $\rvy$, in other words, \textit{good representation might not need disentangled factors}. In theory, it is desirable to learn a complete factorized representation that each dimension individually associates with a single disentangled factor. In practice, the ground truth factors are often entangled, for instance color can be both represented in the RGB space or HSV space, or one might observe the shadow instead of the shape of an object. Moreover, the ground truth factors aren't necessary  the best representations, e.g. wasting two dimensions modeling x and y-axis for $28 \times 28$ images is less efficient than having a single dimension store all $784$ pixels. As a result, learning an independent mechanism that recombines and reuses representation is more robust to the distribution shift in the context of deep generative model~\cite{scholkopf21_causalrepresentationlearning,montero21_roledisentanglementgeneralisation,trauble20_Independenceallyou}. Since study in~\cite{locatello19_ChallengingCommonAssumptions} proves learning such mechanism in an unsupervised manner is infeasible, we focus on the semi-supervised setting which also enables our representation to be controllable by meaningful factors.

\subsection{Semi-supervised maximizing mutual information VAE}
\label{subsec:semafo_vae}

We observe that certain VAE models have high capacity latents, but they generate meaningless images that seems to be the mixture of fine details from multiple training examples. This is particularly common phenomenon when reducing the strength of regularization term in ELBO even though the reconstructed image is much sharper and a t-SNE plot of the latent space shows strong correlation between the latent codes and the ground truth factors\footnote{Additional experiments in the Appendix}. We attribute two explanations to the issue: 1) the lack of hierarchy in the representation due to assumed mean field approximation, all representations are learned equally so that there are large amounts of generative combination; 2) the uninformative prior induces a gap between inference and generation which also render the generation uncontrollable.
In \cite{bozkurt_RateRegularizationGeneralizationVAEs}, the regularization term is decomposed into $\infoq{x}{z} + \dkl{q(\rvz)}{p(\rvz)}$ where $q(\rvz)$ is the aggregated posterior, the authors show that generalization keep improved despite $\infoq{x}{z}$ saturated to the maximum value which indicates the importance of ``the marginal KL'' - $\dkl{q(\rvz)}{p(\rvz)}$ and the choice of more informative prior.

\begin{figure}[ht]
\vspace{-5pt}
\begin{small}
  \begin{center}
    \includegraphics[width=\linewidth]{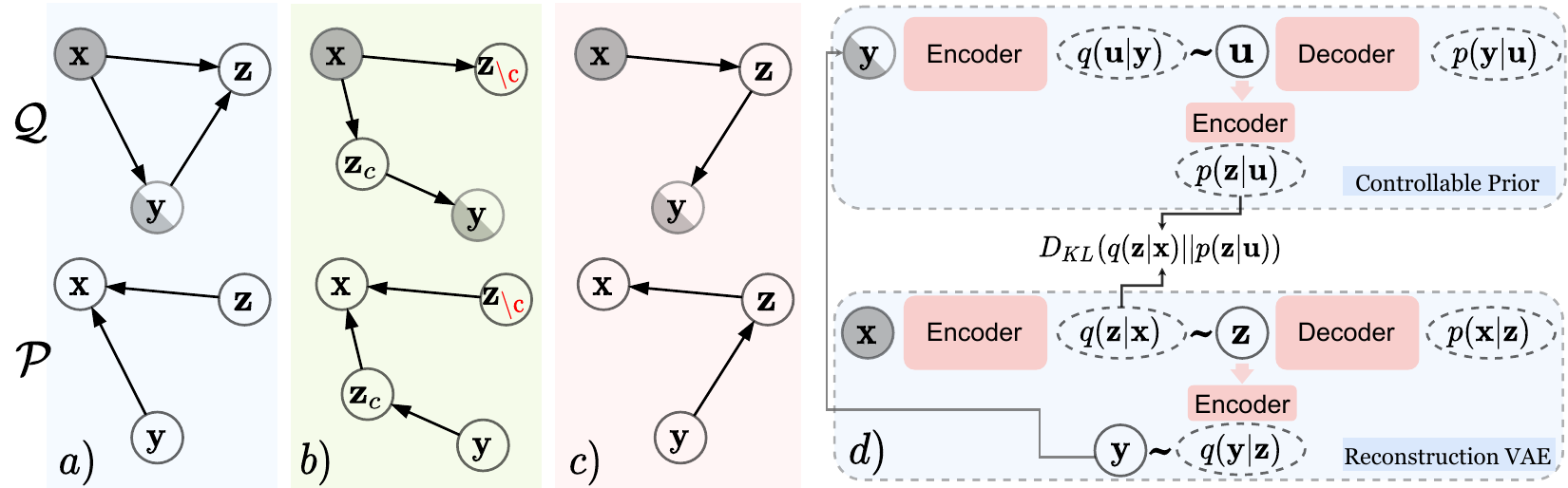}
  \end{center}
  \caption{The graphical models of semi-supervised VAE systems where $\mathcal{P}$ denotes generative model, $\mathcal{Q}$ denotes inference model. $a)$ conditional M2 VAE \cite{kingma14_Semisupervisedlearningdeep}; $b)$ CCVAE \cite{joy21_Capturinglabelcharacteristics}; and $c)$ the proposed SemafoVAE. $d)$ is the implementation of SemafoVAE which consist of two VAEs learned jointly.}
  \label{fig:graphical_model}
\end{small}
\vspace{-0pt}
\end{figure}

As a result, we propose SEmi-supervised MAximizing mutual inFOrmation VAE (SemafoVAE) to \textit{learn meaningful and controllable prior in semi-supervised manner}. Furthermore, the approach is inspired by the ``semaphore'' concept in computer science, that the information bottleneck in VAE is overcome by an alternative pathway created in the prior that maximize $\info{y}{z}$. Figure~\ref{fig:graphical_model}-c shows the graphical model of our approach in comparison to the conditional M2-VAE \cite{kingma14_Semisupervisedlearningdeep} and {\em capturing characteristic VAE} (CCVAE) \cite{joy21_Capturinglabelcharacteristics}. Major differences are the assumption regarding the role of partially observed ground truth $\rvy$ in generating observation $\rvx$, whereas  \cite{kingma14_Semisupervisedlearningdeep} (Figure~\ref{fig:graphical_model}-a) requires $\rvy$ to be marginalized in the generation and the role of $\rvy$ is completely detached from the learned representation. In~\cite{joy21_Capturinglabelcharacteristics} (Figure~\ref{fig:graphical_model}-b), the assumption is that the label characteristics should be captured independently and in parallel with the latent style variables. Lastly, SemafoVAE emphasizes the absolute control of the ground truth factors on the latent space, hence, forcing the representation to be a smooth universal transformation among all classes (e.g. the same mechanism should be used to rotate an image of number 0, 1, or 2).

\subsection{Learning objectives and theoretical justification}

In order to construct an objective for the model above, we formulate a lower bound on the model log-likelihood which factors over the supervised $\mathcal{S}$ subset and unsupervised $\mathcal{U}$ subset of data, i.e.
$p(\rvx,\rvy)=\prod_{\rvx \in \mathcal{U}} p(\rvx) \cdot \prod_{(\rvx,\rvy) \in \mathcal{S}} p(\rvx,\rvy)$.
A detailed derivation for the following ELBOs is in the Appendix~\ref{app:subsec:semafo_derivation}, the objective for unsupervised learning and supervised learning are:
\begin{equation}
\log p(\rvx) \ge
  \expect{q(\rvz|\rvx)}{\log p(\rvx|\rvz)} -
    \expect{ q(\rvy|\rvz) }{ \dkl{q(\rvz|\rvx)}{p(\rvz|\rvy)} } -
    \expect{ q(\rvz|\rvx) }{ \dkl{q(\rvy|\rvz)}{p(\rvy)} }
\label{eq:elbo_uns}
\end{equation}
and
\begin{equation}
  \log p(\rvx,\rvy) \ge \expect{q(\rvz|\rvx)}{\log p(\rvx|\rvz)} -
          \dkl{q(\rvz|\rvx)}{p(\rvz|\rvy)} +
          \log p(\rvy).
\label{eq:elbo_sup}
\end{equation}
Unlike the approach in \cite{kingma14_Semisupervisedlearningdeep,joy21_Capturinglabelcharacteristics} which assume that the prior $p(\rvy)$ is uninformative, we take a more generalized approach. First, $\rvy$ could take any arbitrary distribution, and second we want its prior distribution to be informative and learnable. As a result, we assume that these factors are independently distributed as the factorized joint distribution: $p(\rvy) = \prod_{y \in \mathcal{Y}}p(y)$ where $\mathcal{Y}$ is our set of partially observed factors.

To this point, maximizing $\log p(\rvy)$ is intractable for all possible outcomes in $\mathcal{Y}$, hence, we utilize additional latent variables $\rvu$ in order to maximize the ELBO of $\log p(\rvy)$. Since $\rvy$ has smaller number of dimensions, and its true manifold dimension is the number of disentangled factors, we set the number of latent dimensions $d_\rvu = |\mathcal{Y}|$. ELBO is then,
\begin{equation}
\begin{aligned}
  \log p(\rvy) \ge \expect{q(\rvu|\rvy)}{\log p(\rvy|\rvu)} - \dkl{q(\rvu|\rvy)}{p(\rvu)} = \elbo(\rvy,\rvu).
\end{aligned}
\label{eq:elbo_yu}
\end{equation}
Next, we want to minimize the quantity $\dkl{q(\rvz|\rvx)}{p(\rvz||\rvy)}$ given~\eqref{eq:elbo_yu}. The principle is in the Theorem~\ref{the:trans_info}, as we use $\rvu$ as an auxiliary (bridge) variable for \textit{transferring} information between $\rvy$ and $\rvz$, thus we assume the factorization: $p(\rvz,\rvy,\rvu)=p(\rvy|\rvu)p(\rvz|\rvu)p(\rvu)$, so that:
\begin{equation}
\begin{aligned}
  \dkl{q(\rvz|\rvx)}{p(\rvz||\rvy)}
    &= \expect{q(\rvz|\rvx)}{\log q(\rvz|\rvx)} -
       \underbrace{
         \expect{q(\rvz|\rvx)}{\log \int_u p(\rvz,\rvy,\rvu) \mathrm{d}u}
        }_{\text{applying a lower bound}} +
       \expect{q(\rvz|\rvx)}{\log p(\rvy)} \\
    &\le \expect{ q(\rvz|\rvx)q(\rvu|\rvy) }{ \dkl{q(\rvz|\rvx)}{p(q(\rvz|\rvu)} } -
       \elbo(\rvy,\rvu) +
       \log p(\rvy),
\end{aligned}
\label{eq:elbo_dkl_z}
\end{equation}
where $\elbo(\rvy,\rvu)$ is defined by~\eqref{eq:elbo_yu}. In other words, we could minimize $\dkl{q(\rvz|\rvx)}{p(\rvz||\rvy)}$ by minimizing its upper bound, and the role of $\rvu$ is \textit{beautifully} justified when we substitute \eqref{eq:elbo_dkl_z} to \eqref{eq:elbo_uns} and \eqref{eq:elbo_sup} which eliminates the need for maximizing the intractable evidence $\log p(\rvy)$ and replaces $\rvy$ by the tractable latent $\rvu$ in all of our KL divergence terms. As a result, the final unsupervised and supervised ELBOs for SemafoVAE are:
\begin{equation}
\begin{aligned}
  \elbou(\rvx,\rvz) &= \expect{q(\rvz|\rvx)}{\log p(\rvx|\rvz)} -
                        \expect{ q(\rvz,\rvy,\rvu|\rvx) }{ \dkl{q(\rvz|\rvx)}{p(\rvz|\rvu)} } + \\
           &\quad\quad  \expect{ q(\rvy|\rvz) }{ \elbo(\rvy,\rvu) } -
                        \expect{ q(\rvz|\rvx) }{ \log q(\rvy|\rvz) }\\
\end{aligned}
\label{eq:elbo_uns_final}
\end{equation}
and
\begin{equation}
\begin{aligned}
  \elbos(\rvx,\rvy,\rvz) &= \expect{q(\rvz|\rvx)}{\log p(\rvx|\rvz)} -
                              \expect{ q(\rvz,\rvu|\rvx,\rvy) }{ \dkl{q(\rvz|\rvx)}{p(\rvz|\rvu)} } +
                              \elbo(\rvy,\rvu). \\
\end{aligned}
\label{eq:elbo_sup_final}
\end{equation}
These two objectives are combined into the final ELBO for optimization $\elbosemafo = \elbou(\rvx,\rvz) + \elbos(\rvx,\rvy,\rvz)$.
\begin{lemma}
  For the factorized joint distribution $p(\rvx,\rvy,\rvz)=p_\theta(\rvx|\rvz)p_\psi(\rvz|\rvy)p(\rvy)$, and the assumed inference model $q(\rvx,\rvy,\rvz)=q_\tau(\rvy|\rvz)q_\phi(\rvz|\rvx)q_\data(\rvx)$. Then, there exists the solution parameters $\{\theta,\psi,\tau,\phi\}$ for SemafoVAE, such that the mutual information between the generated example and the ground truth factor $\infop{x}{y}$ is maximized.
  \label{lem:semafo}
\end{lemma}
The lemma is proved using~\cite{barber03_IMalgorithmvariational} lemma to show that given enough data, optimal optimizer and constrained entropy $\entropyp{\rvz|\rvy} \ge 0$, SemafoVAE maximizes the lower bound of $\infop{x}{z}$ and $\infop{z}{y}$, and according to Theorem~\ref{the:trans_info}, maximizing the lower bound of $\infop{x}{y}$. Full detailed proof is in the Appendix~\ref{app:subsec:semafo_maxinfo}. We expect the objective of $\elbosemafo$ to improve the MI of the generated examples and the ground truth factors, i.e. generating relevant example using the prior. The implementation of SemafoVAE is described Figure~\ref{fig:graphical_model}-d which consists of two VAEs optimized jointly. The reconstruction VAE learns the posterior $q(\rvz|\rvx)$ as in the original VAE framework \cite{kingma14_AutoEncodingVariationalBayes}, while the controller VAE learns the controllable prior $p(\rvz|\rvu)$. The training algorithm for SemafoVAE is specified in Appendix~\ref{app:subsec:semafo_training}. It is also notable that our approach operates on the prior distribution, hence, it could be integrated to any existing VAE model as an extension.

\section{Related works}
\label{sec:related}

Richer connection between $\rvz$ and $\rvx$ would increase the capacity of VAE by allowing the modeling of more complicated factors in the generation process. One possibility is to introduce auxiliary variables $\rva$ that factorizes the approximated posterior distribution into $q(\rvz|\rvx) = \int_a q(\rvz|\rva,\rvx) q(\rva|\rvx)\mathrm{d}a$, as a result, enabling complicated covariance structure in $q(\rvz|\rvx)$ ~\cite{maaloe16_AuxiliaryDeepGenerative}. Similar idea could be found in the hierarchical latent models \cite{sonderby16_Laddervariationalautoencoders,maaloe19_BIVAverydeep,child21_VerydeepVAEs} which stack multiple stochastic units to form hierarchical structures $\qphi(\rvz_L|\rvx) = \int_z \qphi(\rvz_L|\rvz_{<L}, \rvx) \qphi(\rvz_{<L}|\rvx) dz$. The learned representation exhibits multiple levels of abstraction, and it also improves the quality of generated images.

Other directions focus on having more powerful posterior distribution \cite{kingma16_Improvedvariationalinference,davidson19_IncreasingExpressivityHypersphericala} or more accurate prior distribution \cite{chen17_VariationalLossyAutoencoder,tomczak18_VAEVampPrior}. The approximated posterior can be for example used to capture clustering via mixture of Gaussian \cite{nalisnick16_ApproximateInferenceDeep}, or encapsulate geometric patterns via hyper-spherical distribution \cite{davidson19_IncreasingExpressivityHypersphericala}. Optimal choice of prior can be approximated without being overfitted to the training data, the result is an empirical mixture of priors that utilizes mixtures of pseudo-inputs $\{ \rvu_1, ...,\rvu_K\}$~\cite{tomczak18_VAEVampPrior}. However, relying on the pseudo-inputs for prior would introduce unnecessary inductive bias to $\rvz$, which consequently limits its capacity to explore the data manifold. In \cite{kumar18_VariationalInferenceDisentangleda} it was shown that disentangled representation needs disentangled prior, and by placing constraints on the covariance structure, we could push the posterior closer to the disentangled prior.

To the best of our knowledge, learning a complete disentangled factors in unsupervised fashion is infeasible \cite{locatello19_ChallengingCommonAssumptions,montero21_roledisentanglementgeneralisation}. Even though there exist successes in incorporating weak supervision that facilitates disentanglement \cite{shu19_WeaklySupervisedDisentanglement,locatello20_WeaklySupervisedDisentanglementCompromises}, this form of supervision might still be too far reach for real world setting since it requires labels for every example. The goal is to achieve disentanglement and controllability via semi-supervised learning with minimal labeling. Similar approaches in \cite{kingma14_Semisupervisedlearningdeep,maaloe16_AuxiliaryDeepGenerative,joy21_Capturinglabelcharacteristics} aim to learn FOVs (styles) and labeling classes in separation, however, our assumption differs that the discovered factors are given by \textit{prior knowledge} about the classes, thus, enabling multiple levels of hierarchy in the prior distribution.

\section{Experiments and results}
\label{sec:experiments}
\begin{wraptable}{r}{0.5\textwidth}
\vspace{-10pt}
\begin{small}
\caption{The baselines system and their improvement compared to the vanilla VAE: $D$ for disentanglement, $L$ for reconstruction log-likelihood and $C$ for controllability of the latent representation. \textit{UNS} is unsupervised methods and \textit{SSL} is semi-supervised methods.}
\label{tab:baselines}
\begin{center}
\begin{adjustbox}{max width=0.5\textwidth}
\begin{tabular}[c]{llccc}
  \toprule
  \multirow{2}{*}{Group} &  \multirow{2}{*}{Method} & \multicolumn{3}{c}{Improvement} \\ \cmidrule{3-5}
  && $D$ & $L$ & $C$ \\
  \midrule
  UNS & BetaVAE \cite{higgins17_betaVAELearningBasic}  & \checkmark & & \\
                & GammaVAE \cite{rezende18_TamingVAEs}  &  & \checkmark & \\
                & FactorVAE \cite{kim18_Disentanglingfactorising} & \checkmark & & \\
                & HierarchicalVAE~\cite{kingma16_Improvedvariationalinference} &  & \checkmark & \\
  \midrule
  SSL & M2   \cite{kingma14_Semisupervisedlearningdeep} & \checkmark & \checkmark & \\
      & CCVAE \cite{joy21_Capturinglabelcharacteristics} &  & & \checkmark \\
      & SemafoVAE & \checkmark  & \checkmark & \checkmark \\
      & SemafoHVAE & \checkmark  & \checkmark & \checkmark \\
  \bottomrule
\end{tabular}
\end{adjustbox}
\end{center}
\end{small}
\vspace{-10pt}
\end{wraptable}

SemafoVAE is compared to a wide range of different approaches from unsupervised to semi-supervised Table~\ref{tab:baselines}. All methods are our reimplementations, and the performance closely match to the description in the original papers. The exception is GammaVAE which is our modification of~\cite{bozkurt_RateRegularizationGeneralizationVAEs} that places extra weight to the reconstruction term of ELBO to specifically improve the log-likelihood. Since SemafoVAE work well as an extension to any existing VAE, we also introduce the \textit{Semafo-HierarchicalVAE} which is the combination of SemafoVAE prior and hierarchical latent variables model~\cite{kingma16_Improvedvariationalinference}.

We utilize the three standard benchmark datasets: MNIST \cite{mnist}, Fashion MNIST (F-MNIST)\cite{fashionmnist} and Shapes3D \cite{shapes3d}. The percent of labelling examples for semi-supervision scenarios are $0.004$, $0.01$ and $0.1$ corresponding to the three datasets. The multi-class labels in the first two datasets are treated as ground truth factors while the discretized factors are used for Shapes3D. We use the benchmark architectures from \cite{locatello19_ChallengingCommonAssumptions}, a Bernoulli distribution is fitted for each pixel and the latent variables are multivariate diagonal normal, and Gumbel-Softmax \cite{jang16_CategoricalReparameterizationGumbelSoftmax} for parameterizing every individual factor $y$. The experimental details can be found in Appendix.

\subsection{Quantitative evaluation}

The SemafoVAE shows consistent improvement to the baseline methods in all three benchmarks, and it is the only semi-supervised method that improves both the generation and the quality of the representation according to the FID and DCI scores. With only two methods have better test log-likelihood (GammaVAE and HVAE) in certain cases, however, both of these methods achieve much lower FID for generated data and DCI for learned representation. A Semafo prior applied to either the vanilla VAE or the HVAE both show significant improvement. Among the semi-supervised approaches, our methods achieve the highest scores in all benchmarks.

\begin{table}[ht]
\vspace{-10pt}
\caption{From left to right: reconstruction log-likelihood (higher is better), FID (\cite{heusel17_GANstrainedtwo}) of random generated samples (lower is better), and the \textit{disentanglement, informativeness, completeness} (DCI~\cite{eastwood18_frameworkquantitativeevaluation}) score between the learned representation and the ground truth factors (higher is better). All scores are calculated using test set. Since there are no ground truth factors for MNIST and F-MNIST, the given DCI score is the accuracy of downstream classifier. $^\dagger$ denotes hierarchical latent models.}
\label{tab:llk}
\begin{center}
\begin{adjustbox}{max width=\textwidth}
\begin{tabular}[c]{lccccccccccc}
  \toprule
        & \multicolumn{3}{c}{Reconstruction Log-likelihood} &&
          \multicolumn{3}{c}{Fr{\'e}chet Inception Distance} &&
          \multicolumn{3}{c}{DCI score}\\
        \cmidrule{2-4} \cmidrule{6-8} \cmidrule{10-12}
  Model      & MNIST           & F-MNIST          & Shapes3D &&
               MNIST           & F-MNIST          & Shapes3D &&
               MNIST           & F-MNIST          & Shapes3D \\
  \midrule
  \multicolumn{5}{l}{\textbf{Unsupervised methods}} \\
  VAE                 & -71.27          & -222.07          & -3464.40 &&
                         48.24          &  97.52           & 74.57 &&
                         89.80          &  78.90           & 64.82 \\
  BetaVAE             & -142.81         & -258.83          & -3492.65 &&
                         107.33         & 159.69           & \textbf{49.89} &&
                         87.25          &  78.05           & 58.56 \\
  GammaVAE            & -54.95          & \textbf{-206.04} & -3456.99 &&
                         51.31          &  119.43          & 141.33 &&
                         89.75          &  81.30           & 48.89 \\
  FactorVAE           & -79.37          & -226.64          & -3470.09 &&
                         46.87          &  95.63           & 115.24 &&
                         85.55          &  78.15           & 63.11 \\
  HVAE$^\dagger$      & -65.84          & -213.50          & \textbf{-3451.54} &&
                         48.68          &  85.99           & 82.20 &&
                         \textbf{92.65} &  82.50           & 69.03 \\
  \midrule
  \multicolumn{5}{l}{\textbf{Semi-supervised methods}} \\
  M2                  & -83.81          & -231.80          & -3464.79 &&
                         64.50          &  105.80          & 81.63 &&
                         47.25          &  59.60           & 24.42 \\
  CCVAE               & -80.53          & -228.63          & -3473.60 &&
                         47.82          &  117.44          & 115.17 &&
                         91.65          &  81.50           & 80.32 \\
  SemafoVAE           & -56.02          & -212.54          & -3457.03 &&
                         45.57          &  92.28           & 93.63 &&
                         90.30          &  81.40           & 80.88 \\
  SemafoHVAE$^\dagger$& \textbf{-52.96} & -209.07          & -3451.80 &&
                        \textbf{42.97}  & \textbf{70.61}   & 62.36 &&
                         90.60          & \textbf{84.05}   & \textbf{82.62} \\
  \bottomrule
\end{tabular}
\end{adjustbox}
\end{center}
\vspace{-10pt}
\end{table}

\subsection{Controlling and disentangling of the representation}

\begin{figure}[ht]
\vspace{-10pt}
\begin{small}
  \begin{center}
    \includegraphics[width=\linewidth]{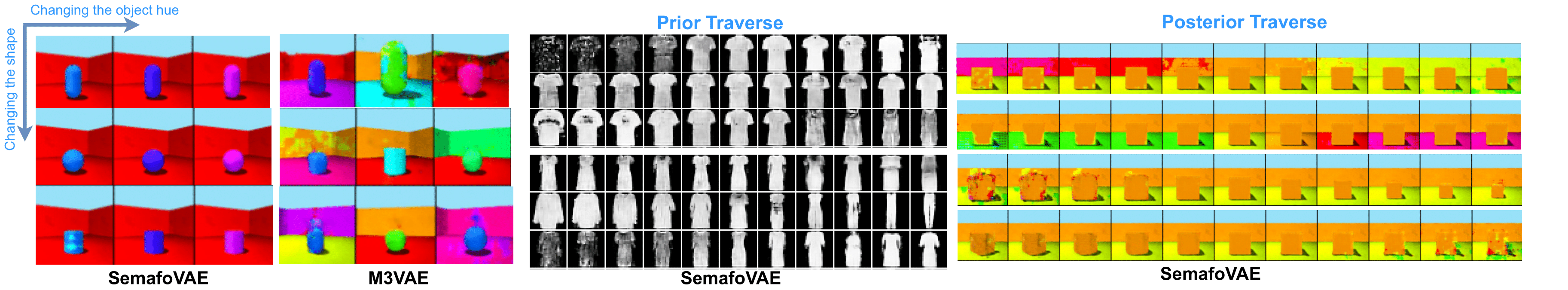}
  \end{center}
  \caption{\textbf{Left}: Random examples generated from the prior conditioned so that only the shapes and the object hue is varied. \textbf{Middle}: Given a shirt (top row) and a dress (bottom row), SemafoVAE learns a conditional prior $p(\rvz|\rvu,\rvy)$ that support smooth traverse in the prior. \textbf{Right}: posterior traverse of a given examples.}
  \label{fig:control_generation}
\end{small}
\vspace{-10pt}
\end{figure}

It is notable that the only existing method that focus on controllability of latent representation is CCVAE \cite{joy21_Capturinglabelcharacteristics}, however, this method only partially control the latent units. Figure~\ref{fig:control_generation} (left) shows that SemafoVAE has perfect control of all factors in generation. Since our method learn an informative prior, it is capable of prior traverse along with the conventional method of posterior traverse. Figure~\ref{fig:control_generation} (middle and right) show that the algorithm has been able to learn smooth traverse of meaningful factors in both of its prior and posterior. The implication is that SemafoVAE can do meaningful intervention of existing attributes, and also supports discovering new combination by traversing completely random prior. More details of the traverse among different VAEs are provided in the Appendix.

\section{Conclusion}
\label{sec:conclusion}

In order to enable practical application, not only generative model has to learn an efficient representation that is hierarchical and disentangled, but also realizes a mechanism that recombines the learned representation in a meaningful way. We present a novel approach that enables VAE to achieve all the proposed criteria. Moreover, we develop the mutual information maximization theory that supports the design of our \textit{Semafo} prior which could be integrated into the existing VAE framework. The prior achieve controllability via adding a minimal amount of supervision using a semi-supervised paradigm. This assumption of variables' hierarchy indirectly introduces inductive bias to the posterior without restricting its ability to explore the factor characteristics. The result is a great degree of controllability while retaining meaningful disentangled latent representation. Since we have successfully integrated SemafoVAE to Hierarchical VAE, our future would focus on scaling up this architecture to achieve controllable realistic image generation.

\bibliographystyle{plainnat}
\bibliography{./references}

\newpage
\appendix


\section{Appendix}
\label{sec:appendix}

\subsection{Proof for Lemma~\ref{lem:elbo_adversarial}: Rethinking the ELBO}
\label{app:subsec:elbo_adversarial}

Given the following factorization: $\qxz(\rvx,\rvz) = \qphi(\rvz|\rvx)\qdata(\rvx)$ and $\pthe(\rvx,\rvz) = \pthe(\rvx|\rvz)p(\rvz)$

By definition in \cite{kingma14_AutoEncodingVariationalBayes}, the ELBO is

\begin{equation}
\begin{aligned}
\elbo(\data,\phi,\theta)
  &= \expectbig{
      \qdata(\rvx) }{
      \expect{\qphi(\rvz|\rvx)}{\loge \pthe(\rvx|\rvz)} - \dkl{\qphi(\rvz|\rvx)}{p(\rvz) }} \\
  &= \expectbig{\qxz(\rvx,\rvz)}{\loge  \frac{\pthe(\rvx|\rvz)p(\rvz)}{\qphi(\rvz|\rvx)} } \\
  &= \expectbig{\qxz(\rvx,\rvz)}{\loge  \frac{\pthe(\rvx,\rvz)\qdata(\rvx)}{\qphi(\rvz|\rvx)\qdata(\rvx)} } \\
  &= \expectbig{\qxz(\rvx,\rvz)}{
        \loge \frac{\pthe(\rvx,\rvz)\qxz(\rvz)\qdata(\rvx)}{\qphi(\rvx,\rvz)\qxz(\rvz)\qdata(\rvx)} }  +
      \expect{\qdata(\rvx)}{\loge \qdata(\rvx)}\\
  &= \expectbig{\qxz(\rvx,\rvz)}{\loge \frac{\pthe(\rvx,\rvz)}{\qxz(\rvz)\qdata(\rvx)} }  -
      \expectbig{\qxz(\rvx,\rvz)}{\loge \frac{\qphi(\rvx,\rvz)}{\qxz(\rvz)\qdata(\rvx)} }  -
      \entropyq{x}
\end{aligned}
\label{app:eq:elbo}
\end{equation}
where the aggregated posterior $\qxz(\rvz)=\int_x \qphi(\rvz|\rvx)\qdata(x) \mathrm{d}x$

Then, the loss function of our VAE is:

\begin{equation}
\begin{aligned}
  \loss(\data,\phi,\theta)
  &= -\elbo(\data,\phi,\theta) \\
  &= \expectbig{\qxz(\rvx,\rvz)}{\loge \frac{\qphi(\rvx,\rvz)}{\qxz(\rvz)\qdata(x)} }  -
      \expectbig{\qxz(\rvx,\rvz)}{\loge \frac{\pthe(\rvx,\rvz)}{\qxz(\rvz)\qdata(x)} } -
      \expect{\qdata(\rvx)}{\loge \qdata(\rvx)}\\
  &= \infoq{x}{z}  -
      \expectbig{\qxz(\rvx,\rvz)}{\loge \frac{\pthe(\rvx,\rvz)}{\qxz(\rvz)\qdata(x)} } +
      \entropyq{x}\\
\end{aligned}
\label{app:eq:elbo_adversarial}
\end{equation}

As for the second term, we want to minimize the loss $\loss(\data,\phi,\theta)$, hence, we are interested in the maximization of the second term and its implication.

\begin{lemma}[\cite{barber03_IMalgorithmvariational} (BA)]
The variational lower bound on mutual information for any set of two random variables $\rvx$ and $\rvy$ is
\begin{displaymath}
  \info{x}{y} \ge \expect{p(\rvx,\rvy)}{\loge q(\rvx|\rvy) - \loge p(\rvx)}
\end{displaymath}
where $q(\rvx|\rvy)$ is an arbitrary variational distribution.
\label{lem:ba}
\end{lemma}

Applying the BA lemma to estimate the lower bound of $\infoq{x}{z}$, so that:

\begin{equation}
\begin{aligned}
  \infoq{x}{z} \ge& \expect{\qxz(\rvx,\rvz)}{\loge \pthe(\rvx|\rvz) - \loge \qdata(\rvx)} \\
      &= \expectbig{\qxz(\rvx,\rvz)}{\loge \frac{\pthe(\rvx|\rvz)p(\rvz)\qxz(\rvz)}{\qdata(\rvx)p(\rvz)\qxz(\rvz)} } \\
      &= \expectbig{\qxz(\rvx,\rvz)}{\loge \frac{\pthe(\rvx,\rvz)}{\qdata(\rvx)\qxz(\rvz)} +
                                     \loge \frac{\qxz(\rvz)}{p(\rvz)}} \\
      &= \expectbig{\qxz(\rvx,\rvz)}{\loge \frac{\pthe(\rvx,\rvz)}{\qdata(\rvx)\qxz(\rvz)}} + \dkl{\qxz(\rvz)}{p(\rvz)} \\
\end{aligned}
\label{app:eq:elbo_ba}
\end{equation}

Since $\dkl{\qxz(\rvz)}{p(\rvz)} \ge 0$ for all $z$, thus:

\begin{equation}
  \infoq{x}{z} \ge \expectbig{\qxz(\rvx,\rvz)}{\loge \frac{\pthe(\rvx,\rvz)}{\qdata(\rvx)\qphi(\rvz)}}
\end{equation}

As stated in the paper, the second term in~\eqref{app:eq:elbo_adversarial} is the lower bound of the encoder's mutual information $\infoq{x}{z}$, as a result, the maximization of ELBO pushes the decoder to recover the encoder's mutual information. Our statements in Lemma~\ref{lem:elbo_adversarial} are proven.

\subsection{Additional Experiments: Limitation of the VAE Framework}
\label{app:subsec:limit_autoencoder}

\textbf{Model Definition}. In order to understand the interaction between the encoder and the decoder in ELBO, we define a more flexible family of BetaVAE~\cite{higgins17_betaVAELearningBasic}, \textit{the BetaGammaVAE}, which introduces scale coefficients to control the influence of both the reconstruction and regularization terms. By varying $\gamma$ and $\beta$ in~\eqref{app:eq:gamma_beta_vae}, we control the lower bound and upper bound of the mutual information $\infoq{x}{z}$.

\begin{equation}
\begin{aligned}
\elbo(\rvx,\rvz;\phi,\theta) =
  \gamma \expect{\qphi(\rvz|\rvx)}{\loge \pthe(\rvx|\rvz)} -
  \beta \dkl{\qphi(\rvz|\rvx)}{p(\rvz)}
\end{aligned}
\label{app:eq:gamma_beta_vae}
\end{equation}

In~\cite{kingma16_Improvedvariationalinference}, \textit{free bits} is the constraint on the minimum amount of information per latents so that the latent units don't collapse to its uninformative prior. Based on the same idea, we proposed the \textit{free pixels} VAE that constrains the maximum amount of likelihood for every individual pixel.
\begin{equation}
\begin{aligned}
\elbo(\rvx,\rvz;\phi,\theta) =
  \sum_{x_i \in \rvx} \bigg( \expect{\qphi(\rvz|x_i)}{\loge \pthe(x_i|\rvz)} + \mathsf{R} \bigg) - \dkl{\qphi(\rvz|\rvx)}{p(\rvz)}
\end{aligned}
\label{app:eq:free_pixels_vae}
\end{equation}
where $\mathsf{R}$ is a non-negative constant. Ideally, higher value of $\mathcal{R}$ would drive the decoder attention to highly informative regions of the image.

In the subsequent paragraphs, we propose the hypotheses that support our claims in Section~\ref{sec:background} and showcase the experiments that corroborate our observation.

\begin{figure}[ht]
\begin{small}
\begin{center}
  \includegraphics[width=\linewidth]{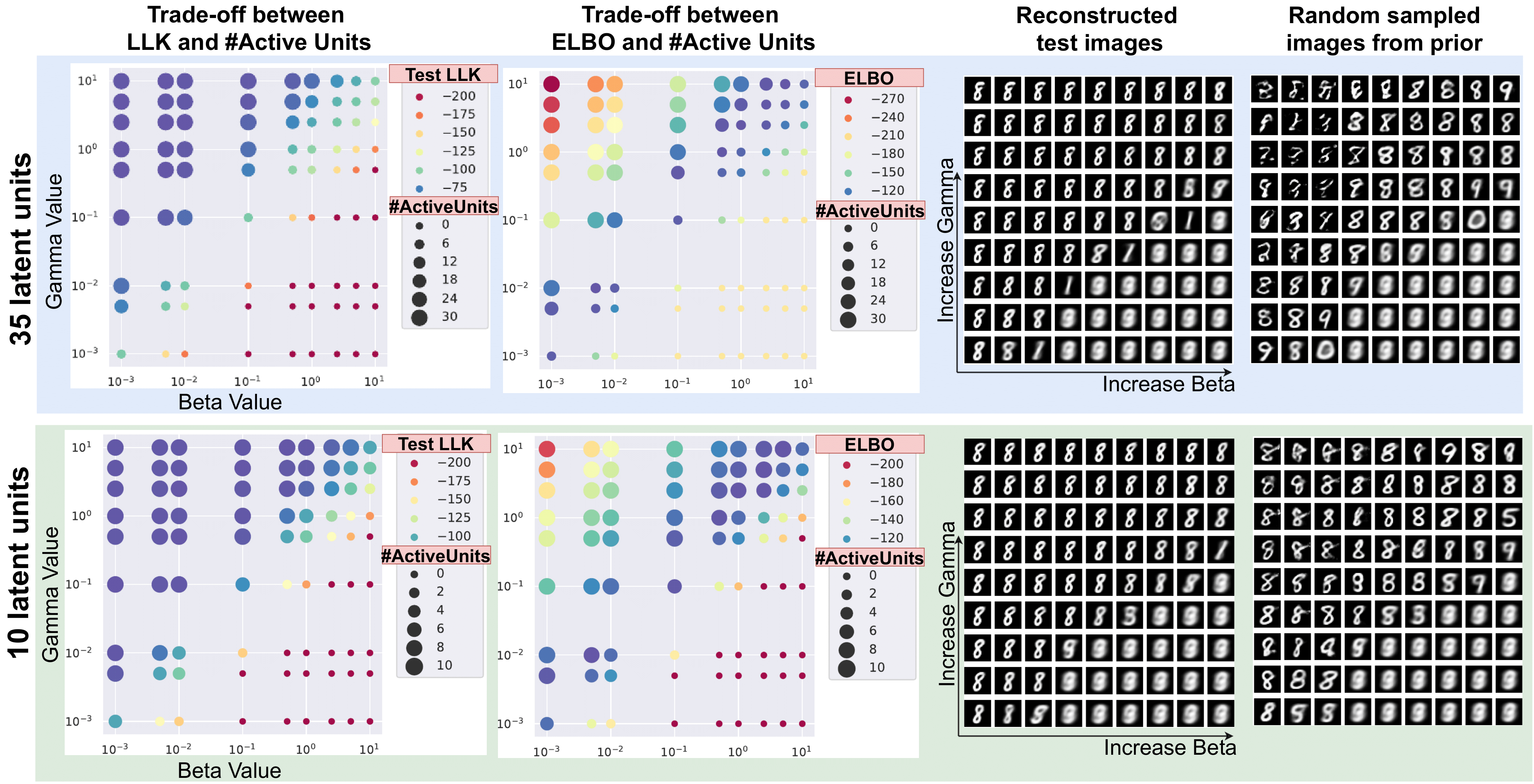}
\end{center}
\caption{Another way to view the rate-distortion trade-off in VAE. Top row shows the trade-off for a VAE with 35 latent units while the bottom row is a VAE with 10 latent units. \textit{Test LLK} is the log-likelihood on test set of a VAE with given $(\gamma,\beta)$ values (higher value is desirable, thus the blue colored dots are better models). \textit{ELBO} is the ELBO on test set (blue dots are higher values). \textit{\#Active Units} is the number of active latent units, i.e. the units don't collapse to their prior. The position of the images is coordinated to the position of the dots, i.e. image and dot in the same row and column have the same $(\gamma,\beta)$ values. \textit{Same network architectures and training configurations were used for all runs}.}
\label{app:fig:rate_distortion}
\end{small}
\end{figure}

\textbf{Hypothesis 1: The ELBO ``game'' must involve two players (encoder and decoder) play optimally} (Section~\ref{subsec:limit_elbo}). Figure~\ref{app:fig:rate_distortion} shows that equal values for $(\gamma,\beta)$ (the bottom left - top right diagonal line) give the best ELBO as well as the best quality of reconstructed test image and random sampled images. Increasing $\gamma$ as in~\cite{bozkurt_RateRegularizationGeneralizationVAEs} has better log-likelihood but disappointing sampling quality, while increasing $\beta$ as in~\cite{higgins17_betaVAELearningBasic} causes the encoder to ignore all the image details and results the worst performance in all the benchmarks.

\textbf{Hypothesis 2: It is hard to control the generation in VAEs with high capacity latents} (Section~\ref{sec:method}). Increasing $\gamma$ enables VAE to capture more image details which increase the latent capacity and the number of active latent units (Figure~\ref{app:fig:rate_distortion}). In practice, this would be the simplest and most effective solution for the posterior collapse issue in VAE. While the random sampled images from high-$\gamma$ VAE with 10 latent dimensions does resemble number ``8'' (Figure~\ref{app:fig:rate_distortion} bottom rightmost figure), VAE with 35 latent dimensions generates a mixed pattern of number ``8''. However, the mixture of patterns includes more details (the line thickness, orientation, and ratio). Hence, we suspect that \textit{the extra activated units coupled with more details make it more difficult for the decoder to search through the meaningful combination of latent representation}.

\begin{figure}[ht]
\begin{small}
\begin{center}
  \includegraphics[width=0.8\linewidth]{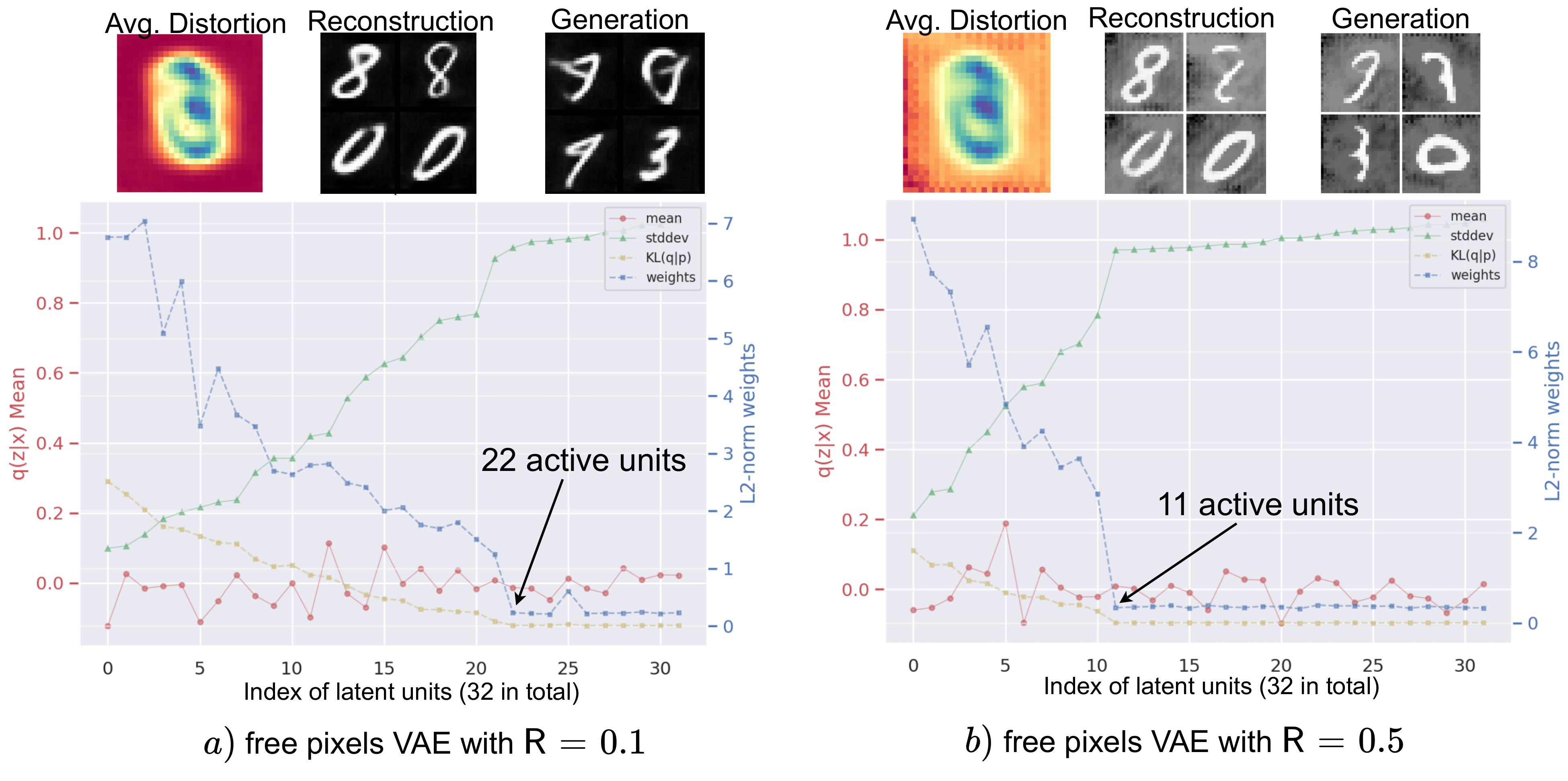}
\end{center}
\caption{Statistics of latent units for VAEs with different values for the ``free likelihood'' constant $\mathsf{R}$.}
\label{app:fig:free_pixel}
\end{small}
\end{figure}

\textbf{Hypothesis 3: MLE decoder doesn't facilitate extrapolation} (Section~\ref{subsec:limit_mle}). Figure~\ref{app:fig:free_pixel} shows that simply constrain the upper bound of likelihood won't resolve the issue with the learning mechanism of MLE. The MLE learner doesn't focus on the high detail (low density) region until all the high-density regions are optimized. This is due to the direction of maximum likelihood learning takes the form of $\dkl{p(\rvx)}{q(\rvx)}$ \cite{bishop06_Patternrecognitionmachine}, any area with high likelihood (the empty region in MNIST image) receives significant more weight than high information details that only occasionally appeared in the image.

\begin{figure}[ht]
\begin{small}
\begin{center}
  \includegraphics[width=\linewidth]{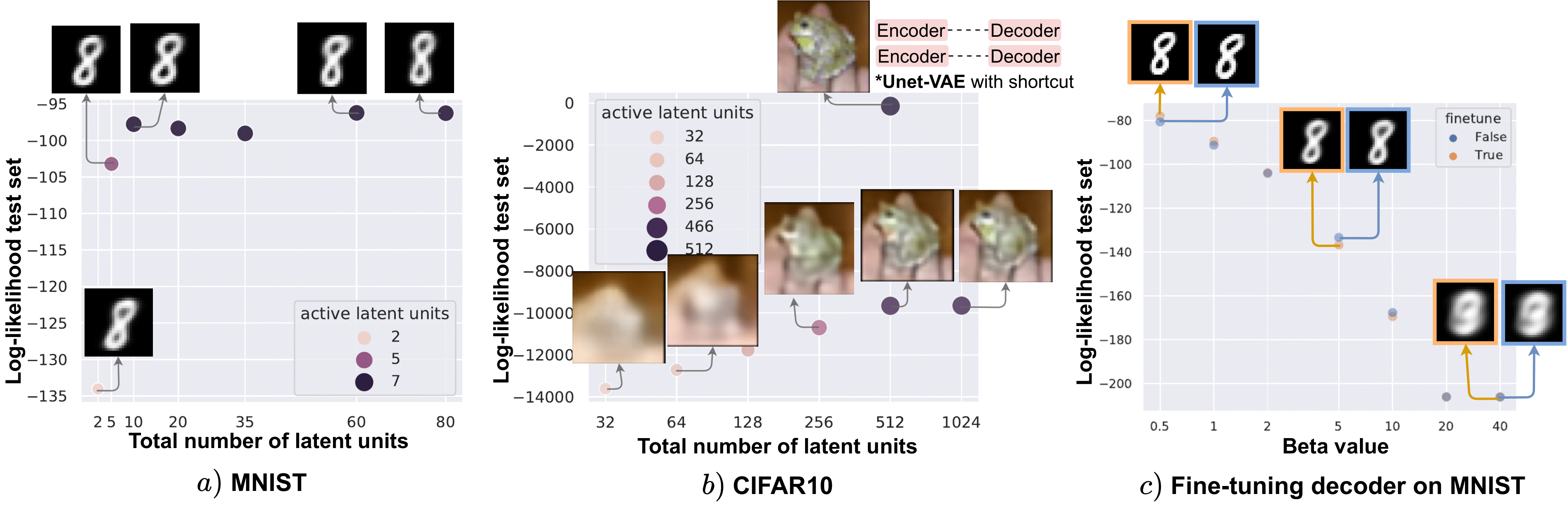}
\end{center}
\caption{In $a)$ and $b)$, we vary the total number of latent units given the same network architectures, the networks trained on MNIST and CIFAR10 respectively. The dot size indicates how many latent units are active. \textit{Unet-VAE} is a vanilla VAE with shortcut connection between every respectively encoder and decoder layers. $c)$ show the same VAE trained with different $\beta$ value. The blue dots are the models trained for 200000 iterations. The orange dots are the models trained for 100000 iterations, then the encoder's weights are fixed, and a new decoder is ``plug-in'' and trained for another 100000 iterations.}
\label{app:fig:active_vae}
\end{small}
\end{figure}

\textbf{Hypothesis 4: VAE converges to the same number of active units regardless of the total number of latent units} (Section~\ref{subsec:limit_ae}). Figure~\ref{app:fig:active_vae}-a shows that for the given network architectures the VAE could only utilize 7 latent units at maximum. Given less than 7 latent units, an implicit constraint is placed for the encoder that the image details are reduced (i.e. blurry image) even though all latent units are activated. In contrast, the VAE decoder ignores all redundant units if more than 7 latent units are given. This indicates that the given network architectures have an upper bound for how much information of observation it could learn and pass through the bottleneck. The same phenomenon applied for CIFAR10 dataset (Figure~\ref{app:fig:active_vae}-b), however, we could clearly observe how much more details the VAE is capable of capturing given more latent units.

\textbf{Hypothesis 5: The autoencoder design prevents fine detail information reaching the bottleneck} (Section~\ref{subsec:limit_ae}). To understand if the ELBO or the autoencoder design imposes an upper limit on the test log-likelihood of VAE, we propose \textit{Unet-VAE} (Figure~\ref{app:fig:active_vae}-b) which introduces shortcut connection between every encoder and decoder layer respectively. Ultimately, this causes the network to completely ignore the latent in the bottleneck. However, a significant amount of fine details including the texture of the frog skin are recovered in the reconstructed image. We suspect that the autoencoder design doesn't allow low-level features (i.e. the fine details) learned from the lower layer to reach the bottleneck, thus the design of Hierarchical VAE tackles this issue and achieved great success in generating realistic images \cite{kingma16_Improvedvariationalinference,child21_VerydeepVAEs}.

\textbf{Hypothesis 6: The encoder imposes an upper bound to the mutual information of the decoder} (Section~\ref{subsec:limit_ae}). Figure~\ref{app:fig:active_vae}-c shows the difference between a VAE with and without a fine-tuned decoder. In conclusion, fine-tuning the decoder doesn't improve the log-likelihood on the test set or the reconstruction quality of the test image. Notably, the same observation is repeated for all VAE with different latent capacity, i.e. different $\beta$ values.

Let denote the information stored in the latents of a VAE is $\info{x}{z}$, the fine-tuned decoder learn to reconstruct image $\mathbf{\hat{x}}$ from the fixed latent $\rvz$, then we have three variables form the Markov chain $\rvx \to \rvz \to \mathbf{\hat{x}}$, and according to the \textit{data processing inequality}~\cite{cover06_Elementsinformationtheory}, $\info{x}{z} \ge \info{\hat{x}}{z}$. In other words, \textit{the pretrained encoder places an upper bound on the mutual information of the fine-tuned decoder, proof for the Proposition~\ref{prop:decoder_bound}}.

A total of \textit{701 experiments} on a single GTX 1080 GPU have been run for this section.

\subsection{Proof of Theorem~\ref{the:trans_info}: Transitive Information}
\label{app:subsec:trans_info}

\begin{proof}
The following properties of entropy is true for any given set of three random variables $\rvx$, $\rvy$ and $\rvz$:

\begin{itemize}
  \item $\info{x}{z} = \entropy{x} - \entropy{x|z}$, similar derivations for $\info{x}{y}$ and $\info{y}{z}$
  \item $\entropy{x,y} = \entropy{x|y} + \entropy{y}$, and $\entropy{x,y|z} = \entropy{x|y,z} + \entropy{y|z}$
  \item $\entropy{x|z} \le \entropy{x,y|z}$
  \item $\entropy{x|y,z} \le \entropy{x|y}$
\end{itemize}

We have:
\begin{align*}
     & \entropy{x|z}               &\le&\quad\quad  \entropy{x,y|z} \\
\iff & \entropy{x|z}               &\le&\quad\quad  \entropy{x|y,z} + \entropy{y|z} \\
\iff & \entropy{x|z}               &\le&\quad\quad  \entropy{x|y} + \entropy{y|z} \\
\iff & \entropy{x} - \entropy{x|z} &\ge&\quad\quad  \entropy{x} - \entropy{x|y} + \entropy{y} - \entropy{y|z} - \entropy{y} \\
\iff & \info{x}{z}                 &\ge&\quad\quad  \info{x}{y} + \info{y}{z} - \entropy{y} \\
\end{align*}
\end{proof}

As a result, Theorem~\ref{the:trans_info} is proven. It is notable that the proof share the same approach as the variation of information metric in~\cite{meila03_Comparingclusteringsvariation}.

\newpage
\subsection{ELBOs derivation for SemafoVAE}
\label{app:subsec:semafo_derivation}

According to Figure~\ref{fig:graphical_model}-c, the modeling assumption of SemafoVAE are:
\begin{itemize}
  \item $p(\rvx,\rvy,\rvz) = p(\rvx|\rvz)p(\rvz|\rvy)p(\rvy)$ for the generative model, and
  \item $q(\rvx,\rvy,\rvz) = q(\rvx)q(\rvz|\rvx)q(\rvy|\rvz)$ for the inference model (\textit{the parameters are omitted in all derivations within this section}).
\end{itemize}

\textbf{ELBO derivation for unsupervised learning}

\begin{align*}
  \log p(\rvx) &= \log \int_{z,y} p(\rvx,\rvy,\rvz) \mathrm{d}z \mathrm{d}y \\
    &= \log \int_z \frac{p(\rvx,\rvy,\rvz)}{q(\rvz,\rvy|\rvx)}q(\rvz,\rvy|\rvx) \mathrm{d}z \\
    &= \log \expectbig{q(\rvz,\rvy|\rvx)}{
         \frac{ p(\rvx|\rvz)p(\rvz|\rvy)p(\rvy) }{ q(\rvz|\rvx)q(\rvy|\rvz) }
        }
\end{align*}
Applying Jensen's inequality:
\begin{equation}
\begin{aligned}
  \loge p(\rvx)
  &\ge \expectbig{ q(\rvz|\rvx)q(\rvy|\rvz) }{ \loge \frac{p(\rvx|\rvz)p(\rvz|\rvy)p(\rvy)}{q(\rvz|\rvx)q(\rvy|\rvz)} } \\
  & =  \expect{ q(\rvz|\rvx) }{ \loge p(\rvx|\rvz) } -
    \expectbig{ q(\rvz|\rvx)q(\rvy|\rvz) }{ \loge \frac{q(\rvz|\rvx)}{p(\rvz|\rvy)} } -
    \expectbig{ q(\rvz|\rvx)q(\rvy|\rvz) }{ \loge \frac{q(\rvy|\rvz)}{p(\rvy)} } \\
  & = \expect{q(\rvz|\rvx)}{\loge p(\rvx|\rvz)} -
    \expect{ q(\rvy|\rvz) }{ \dkl{q(\rvz|\rvx)}{p(\rvz|\rvy)} } -
    \expect{ q(\rvz|\rvx) }{ \dkl{q(\rvy|\rvz)}{p(\rvy)} } \\
  & = \elbou(\rvx,\rvz)
\end{aligned}
\label{app:eq:semafo_uns}
\end{equation}

\textbf{ELBO derivation for supervised learning:}
\begin{equation}
\begin{aligned}
  \log p(\rvx,\rvy)
  &=   \loge \int_{z} p(\rvx,\rvy,\rvz) \mathrm{d}z \\
  &=   \loge \expectbig{ q(\rvz|\rvx) }{ \frac{p(\rvx|\rvz)p(\rvz|\rvy)p(\rvy)}{q(\rvz|\rvx)} } \\
  &\ge \expectbig{ q(\rvz|\rvx) }{ \loge \frac{p(\rvx|\rvz)p(\rvz|\rvy)p(\rvy)}{q(\rvz|\rvx)} } \\
  &=   \expect{q(\rvz|\rvx)}{\loge p(\rvx|\rvz)} -
          \dkl{q(\rvz|\rvx)}{p(\rvz|\rvy)} +
          \loge p(\rvy) \\
      & = \elbos(\rvx,\rvy,\rvz)
\end{aligned}
\label{app:eq:semafo_sup}
\end{equation}

We introduce the latent variables $\rvu$ to maximize the ELBO of $\log p(\rvy)$:

\begin{equation}
\begin{aligned}
  \loge p(\rvy) \ge \expect{q(\rvu|\rvy)}{\log p(\rvy|\rvu)} - \dkl{q(\rvu|\rvy)}{p(\rvu)} = \elbo(\rvy,\rvu)
\end{aligned}
\label{app:eq:elbo_yu}
\end{equation}

\textbf{We derive the upper bound for $\dkl{q(\rvz|\rvx)}{p(\rvz||\rvy)}$}, assumed the factorization $p(\rvz,\rvy,\rvu)=p(\rvy|\rvu)p(\rvz|\rvu)p(\rvu)$.

\begin{equation}
\begin{aligned}
    &\dkl{q(\rvz|\rvx)}{p(\rvz||\rvy)} \\
    =& \expectbig{ q(\rvz|\rvx) }{ \loge \frac{q(\rvz|\rvx)}{p(\rvz|\rvy)} } \\
    =& \expect{ q(\rvz|\rvx) }{ \loge q(\rvz|\rvx) } -
       \expect{ q(\rvz|\rvx) }{ \loge p(\rvz|\rvy) } \\
    =& \expect{ q(\rvz|\rvx) }{ \loge q(\rvz|\rvx) } -
       \expect{ q(\rvz|\rvx) }{ \loge p(\rvz,\rvy) } +
       \expect{ q(\rvz|\rvx) }{ \loge p(\rvy) } \\
    =& \expect{q(\rvz|\rvx)}{\loge q(\rvz|\rvx)} -
       \expect{q(\rvz|\rvx)}{\loge \int_u p(\rvz,\rvy,\rvu) \mathrm{d}u} +
       \expect{q(\rvz|\rvx)}{\loge p(\rvy)} \\
    =& \expect{q(\rvz|\rvx)}{\loge q(\rvz|\rvx)} -
       \underbrace{
          \expect{q(\rvz|\rvx)}{\loge \int_u
               \frac{p(\rvz,\rvy,\rvu)q(\rvu|\rvy)}{q(\rvu|\rvy)} \mathrm{d}u}
        }_{\ge
        \expect{q(\rvz|\rvx)q(\rvu|\rvy)}{\loge
          \frac{p(\rvz,\rvy,\rvu)}{q(\rvu|\rvy)}} \text{(Jensen inequality)}
        } +
        \expect{q(\rvz|\rvx)}{\loge p(\rvy)} \\
    \le& \expect{q(\rvz|\rvx)}{\loge q(\rvz|\rvx)} -
          \expectbig{q(\rvz|\rvx)q(\rvu|\rvy)}{\loge
            \frac{p(\rvy|\rvu)p(\rvz|\rvu)p(\rvu)}{q(\rvu|\rvy)}} +
          \expect{q(\rvz|\rvx)}{\loge p(\rvy)} \\
    =& \expect{ q(\rvz|\rvx)q(\rvu|\rvy) }{ \dkl{q(\rvz|\rvx)}{p(\rvz|\rvu)} } - \elbo(\rvy,\rvu) + \log p(\rvy)
\end{aligned}
\label{app:eq:semafo_dkl_z}
\end{equation}
where $\elbo(\rvy,\rvu)$ is defined by~\eqref{app:eq:elbo_yu}. Finally, we substitute \eqref{app:eq:semafo_dkl_z} to \eqref{app:eq:semafo_uns} and \eqref{app:eq:semafo_sup} which eliminates the need for maximizing the intractable evidence $\loge p(\rvy)$ and replace $\rvy$ by the tractable latents $\rvu$ in all of our KL divergence terms.

The final unsupervised and supervised ELBO for SemafoVAE are:

\begin{equation}
\begin{aligned}
  \elbou(\rvx,\rvz)
  &= \expect{q(\rvz|\rvx)}{\loge p(\rvx|\rvz)} -
     \expect{ q(\rvy|\rvz) }{ \dkl{q(\rvz|\rvx}{p(\rvz|\rvy)} } -
     \expect{ q(\rvz|\rvx) }{ \dkl{q(\rvy|\rvz)}{p(\rvy)} } \\
  &\ge \expect{q(\rvz|\rvx)}{\loge p(\rvx|\rvz)} - \\
        & \quad\quad
        \expectbig{q(\rvy|\rvz)}{
          \expect{ q(\rvz|\rvx)q(\rvu|\rvy) }{ \dkl{q(\rvz|\rvx)}{p(q(\rvz|\rvu)} } +
          \elbo(\rvy,\rvu) -
          \log p(\rvy)
        } - \\
        & \quad\quad
        \bigg( \expect{ q(\rvz|\rvx) }{ \loge q(\rvy|\rvz) } -
               \expect{ q(\rvy|\rvz) }{ \loge p(\rvy) } \bigg) \\
  &= \expect{q(\rvz|\rvx)}{\loge p(\rvx|\rvz)} -
     \expect{ q(\rvz,\rvy,\rvu|\rvx) }{ \dkl{q(\rvz|\rvx)}{p(\rvz|\rvu)} } +
     \expect{ q(\rvy|\rvz) }{ \elbo(\rvy,\rvu) } - \\
     & \quad\quad
     \expect{ q(\rvz|\rvx) }{ \loge q(\rvy|\rvz) }\\
\end{aligned}
\label{app:eq:semafo_uns_final}
\end{equation}
and
\begin{equation}
\begin{aligned}
  \elbos(\rvx,\rvy,\rvz)
  &= \expect{q(\rvz|\rvx)}{\loge p(\rvx|\rvz)} -
     \dkl{q(\rvz|\rvx)}{p(\rvz|\rvy)} +
     \loge p(\rvy) \\
  &\ge \expect{q(\rvz|\rvx)}{\log p(\rvx|\rvz)} - \\
       & \quad\quad
       \bigg( \expect{ q(\rvz|\rvx)q(\rvu|\rvy) }{ \dkl{q(\rvz|\rvx)}{p(\rvz|\rvu)} } -
       \elbo(\rvy,\rvu) + \log p(\rvy) \bigg) + \\
       & \quad\quad
       \log p(\rvy) \\
  &= \expect{q(\rvz|\rvx)}{\log p(\rvx|\rvz)} -
       \expect{ q(\rvz,\rvu|\rvx,\rvy) }{ \dkl{q(\rvz|\rvx)}{p(q(\rvz|\rvu)} } +
       \elbo(\rvy,\rvu) \\
\end{aligned}
\label{app:eq:semafo_sup_final}
\end{equation}

\newpage
\subsection{Hierarchical VAE and the Semafo prior}
\label{app:subsec:semafo_hvae}

\begin{wrapfigure}{r}{0.3\textwidth}
\vspace{-10pt}
\begin{small}
  \begin{center}
    \includegraphics[width=\linewidth]{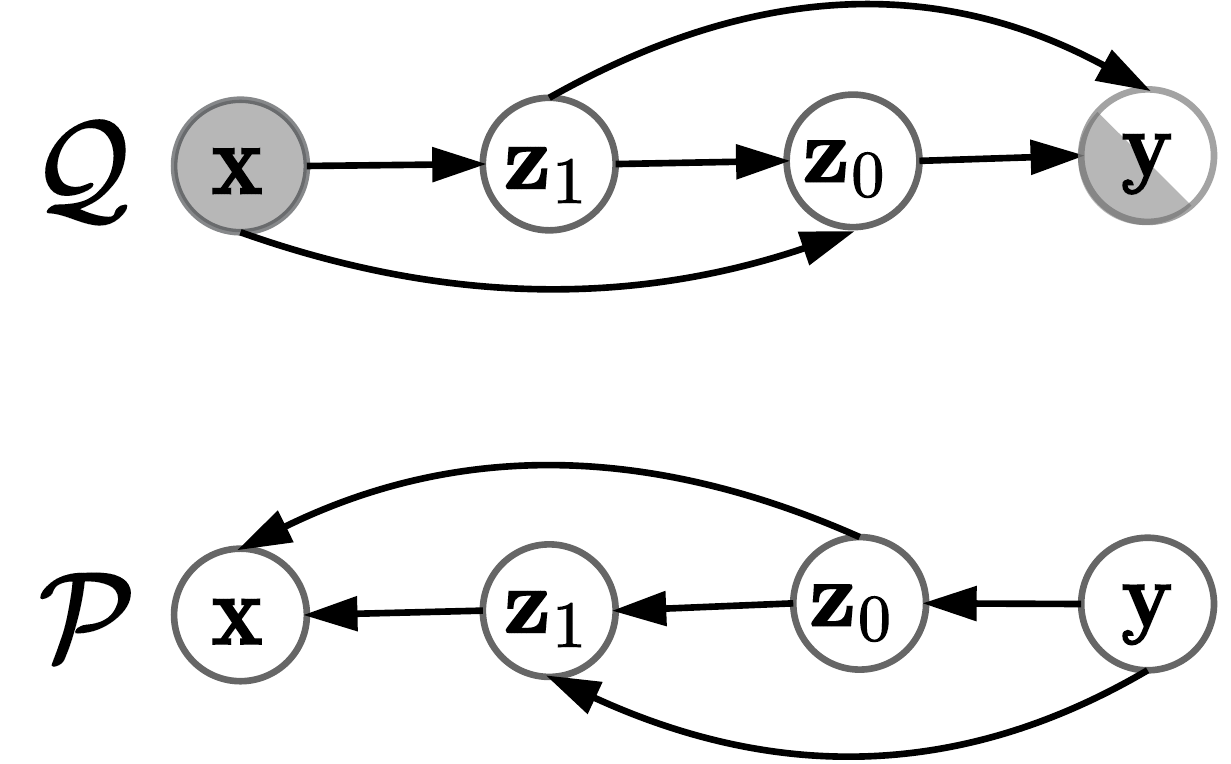}
  \end{center}
  \caption{Graphical model of Semafo Hierarchical VAE with two layers of hierarchical latent variables.}
  \label{app:fig:semafo_hvae}
\end{small}
\vspace{-10pt}
\end{wrapfigure}

In the paper, we also apply the Semafo prior to Hierarchical VAE \cite{kingma16_Improvedvariationalinference,child21_VerydeepVAEs}. According to Figure~\ref{app:fig:semafo_hvae}, a system of $L$-layers hierarchical variables assumes the factorization:
\begin{itemize}
  \item $p(\rvx,\rvz,\rvy)=p(\rvx|\rvz_0, ...,\rvz_L)\: \prod_i^L p(\rvz_i|\rvz_{<i},\rvy)\: p(\rvy)$ and
  \item $q(\rvx,\rvz,\rvy)=q(\rvy|\rvz_0, ...,\rvz_L)\: \prod_i^L q(\rvz_i|\rvz_{<i},\rvx)\: q(\rvx)$
\end{itemize}

Since we only use two layers of the hierarchical latent variables $\rvz_0$ and $\rvz_1$, the following derivation is only for such model, however, the same derivation could be generalized to more latent layers.

\textbf{The unsupervised and supervised ELBO of SemafoHVAE} are

\begin{align}
\elbou(\rvx,\rvz) =
&\; \expect{q(\rvz|\rvx)}{\loge p(\rvx|\rvz)} - \nonumber\\
&\; \expect{q(\rvz_0|\rvz_1,\rvx)q(\rvy|\rvz_0,\rvz_1,\rvx)}{ \dkl{q(\rvz_1|\rvx)}{p(\rvz_1|\rvz_0,\rvy)} } - \nonumber\\
&\; \expect{q(\rvz_1|\rvx)q(\rvy|\rvz_0,\rvz_1,\rvx)}{ \dkl{q(\rvz_0|\rvz_1,\rvx)}{p(\rvz_0|\rvy)} } - \nonumber\\
&\; \expect{q(\rvz_1|\rvx)q(\rvz_0|\rvz_1,\rvx)}{ \dkl{q(\rvy|\rvz_1,\rvz_0,\rvx)}{p(\rvy)} } \label{app:eq:semafoh_uns} \\
\text{and}&\nonumber\\
\elbou(\rvx,\rvy,\rvz) =
&\; \expect{q(\rvz|\rvx)}{\loge p(\rvx|\rvz)} - \nonumber\\
&\; \expect{q(\rvz_0|\rvz_1,\rvx)}{ \dkl{q(\rvz_1|\rvx)}{p(\rvz_1|\rvz_0,\rvy)} } - \nonumber\\
&\; \expect{q(\rvz_1|\rvx)}{ \dkl{q(\rvz_0|\rvz_1,\rvx)}{p(\rvz_0|\rvy)} } + \nonumber\\
&\; \loge p(\rvy) \label{app:eq:semafoh_sup}
\end{align}

\textbf{We expand the KL-terms for $\rvz_0$ and $\rvz_1$}

\begin{align}
\dkl{q(\rvz_0|\rvz_1,\rvx)}{p(\rvz_0|\rvy)}
  &= -\entropyq{\rvz_0|\rvz_1,\rvx} - \expectbig{q(\rvz_0|\rvz_1,\rvx)}{\loge \frac{p(\rvz_0,\rvy)}{p(\rvy)} } \label{app:eq:semafoh_kl_z0}\\
\dkl{q(\rvz_1|\rvx)}{p(\rvz_1|\rvz_0,\rvy)}
  &= -\entropyq{\rvz_1|\rvx} - \expectbig{q(\rvz_1|\rvx)}{\loge \frac{p(\rvz_0,\rvz_1,\rvy)}{p(\rvz_0,\rvy)} } \label{app:eq:semafoh_kl_z1}
\end{align}

Because $\loge p(\rvz_0,\rvy)$ is eliminated when summing \eqref{app:eq:semafoh_kl_z0} and \eqref{app:eq:semafoh_kl_z1}, we focus on the term $\loge p(\rvz_0,\rvz_1,\rvy)$. Applying the strategy as in Section~\ref{app:subsec:semafo_derivation}, introducing the latent variable $\rvu$ to maximize $\loge p(\rvy)$ without assuming a prior $p(\rvy)$ in order to increase the expressiveness of the model and facilitating richer connection between $\rvy$ and $\rvz$.

\textbf{The lower bound of $\expect{q(\rvz_1|\rvx)}{\loge p(\rvz_0,\rvz_1,\rvy)}$} is, assuming the factorization $p(\rvz_0,\rvz_1,\rvy,\rvu)=p(\rvz_1|\rvz_0,\rvu)p(\rvz_0|\rvu)p(\rvy|\rvu)p(\rvu)$,

\begin{equation}
\begin{aligned}
  \expect{q(\rvz_1|\rvx)}{\loge p(\rvz_0,\rvz_1,\rvy)}
    &=   \expectbig{q(\rvz_1|\rvx)}{\loge \int_u p(\rvz_0,\rvz_1,\rvy,\rvu) \mathrm{d}u } \\
    &\ge \expectbig{q(\rvz_1|\rvx)q(\rvu|\rvy)}{\loge \frac{p(\rvz_0,\rvz_1,\rvy,\rvu)}{q(\rvu|\rvy)} } \\
    &= \expect{q(\rvz_1|\rvx)q(\rvu|\rvy)}{\loge p(\rvz_1|\rvz_0,\rvu) } + \loge p(\rvz_0|\rvu) + \elbo(\rvy,\rvu)
\end{aligned}
\label{app:eq:semafoh_kl_upper}
\end{equation}
where $\elbo(\rvy,\rvu)$ is defined in~\eqref{app:eq:elbo_yu}. Now, we substitute~\eqref{app:eq:semafoh_kl_upper} to~\eqref{app:eq:semafoh_kl_z1}, then, substitute~\eqref{app:eq:semafoh_kl_z0} and~\eqref{app:eq:semafoh_kl_z1} to the ELBOs of SemafoHVAE. The two notable outcomes are: \textit{i)} maximizing $\loge p(\rvy)$ is now tractable via $\rvu$; \textit{ii)} all $\rvy$ in the KL-divergence terms of $\rvz$ are replaced by the tractable distribution $\rvu$.

\textbf{The final objectives of SemafoHVAE} are

\begin{align}
  \elbou(\rvx,\rvz) =
  &\; \expect{q(\rvz|\rvx)}{\loge p(\rvx|\rvz)} - \nonumber\\
  &\; \expect{q(\rvz_0|\rvz_1,\rvx)q(\rvy|\rvz_0,\rvz_1,\rvx)q(\rvu|\rvy)}{ \dkl{q(\rvz_1|\rvx)}{p(\rvz_1|\rvz_0,\rvu)} } - \nonumber\\
  &\; \expect{q(\rvz_1|\rvx)q(\rvy|\rvz_0,\rvz_1,\rvx)q(\rvu|\rvy)}{ \dkl{q(\rvz_0|\rvz_1,\rvx)}{p(\rvz_0|\rvu)} } - \nonumber\\
  &\; \expect{q(\rvz_1,\rvz_0,\rvy|\rvx)}{ \elbo(\rvy,\rvu) } + \entropyq{\rvy|\rvz_1,\rvz_0,\rvz_x} \label{app:eq:semafoh_uns_final}\\
  \text{and}&\nonumber\\
  \elbou(\rvx,\rvy,\rvz) =
  &\; \expect{q(\rvz|\rvx)}{\loge p(\rvx|\rvz)} - \nonumber\\
  &\; \expect{q(\rvz_0|\rvz_1,\rvx)q(\rvu|\rvy)}{ \dkl{q(\rvz_1|\rvx)}{p(\rvz_1|\rvz_0,\rvu)} } - \nonumber\\
  &\; \expect{q(\rvz_1|\rvx)q(\rvu|\rvy)}{ \dkl{q(\rvz_0|\rvz_1,\rvx)}{p(\rvz_0|\rvu)} } + \nonumber\\
  &\; \elbo(\rvy,\rvu) \label{app:eq:semafoh_sup_final}
\end{align}

\subsection{Proof Lemma~\ref{lem:semafo}: Theoretical Justification of SemafoVAE}
\label{app:subsec:semafo_maxinfo}

According to the definition in Section~\ref{subsec:semafo_vae}, the SemafoVAE is optimized using data from two subsets and their empirical data distribution: the unsupervised subset $\mathcal{U}$ with $q_\mathcal{U}(x)$, and the supervised subset $\mathcal{S}$ with $q_\mathcal{S}(x, y)$.

In order to understand the ultimate result of optimizing SemafoVAE according to~\eqref{app:eq:semafo_uns_final} and~\eqref{app:eq:semafo_sup_final}, our assumptions are:
\begin{enumerate}[label=(\roman*)]
  \item \textit{unlimited amount of data}: there are enough data in $\mathcal{U}$ and $\mathcal{S}$ so that the empirical distribution $q_\mathcal{U}(\rvx)$ and $q_\mathcal{S}(\rvx,\rvy)$ converge to the actual corresponding data distribution $p(\rvx)$ and $p(\rvx,\rvy)$, i.e. $q_\mathcal{U}(\rvx) \equiv p(\rvx)$ and $q_\mathcal{S}(\rvx,\rvy) \equiv p(\rvx,\rvy)$ ;
  \item \textit{optimal optimization algorithm}: so that the maximum value of ELBOs are realized in both unsupervised and supervised objective.
\end{enumerate}
Now we set to investigate the impact of SemafoVAE on the quantity $\infoq{x}{y}$ (for simplicity we drop dataset subscript in following derivation).

\textbf{For unsupervised case and the empirical distribution $q(\rvx)$}, maximization of ELBO is the equivalent to the maximization of
\begin{equation}
\begin{aligned}
   &\quad \expectbig{q(\rvx)}{\expectbig{ q(\rvz,\rvy|\rvx) }{ \loge \frac{p(\rvx,\rvy,\rvz)}{q(\rvz,\rvy|\rvx)}} - \log p(\rvx)} \\
  =&\quad \expectbig{q(\rvx)}{\expectbig{ q(\rvz,\rvy|\rvx) }{ \loge \frac{p(\rvy,\rvz|\rvx)p(\rvx)}{q(\rvz,\rvy|\rvx)p(\rvx)}}} \\
  =&\quad -\expect{q(\rvx)}{\dkl{q(\rvz,\rvy|\rvx)}{p(\rvz,\rvy|\rvx)}}
\end{aligned}
\label{app:eq:maxinfo_uns}
\end{equation}
As a result, the optimal solution for $\elbou(\rvx,\rvz)$ is $\dkl{q(\rvz,\rvy|\rvx)}{p(\rvz,\rvy|\rvx)}=0$ ($\forall x \sim q(\rvx)$), and because of the unlimited data assumption, hence,
$q(\rvz,\rvy) \equiv p(\rvz,\rvy) ~\refstepcounter{equation}(\theequation)\label{app:eq:maxinfo_uns_yz}$

\textbf{For supervised case and the empirical distribution $q(\rvx,\rvy)$}, we have:
\begin{itemize}
  \item $q(\rvz|\rvx) = q(\rvz|\rvx,\rvy)$ according to the assumed Markov chain $\mathbf{X} \to \mathbf{Z} \to \mathbf{Y}$ of inference model, and
  \item $q(\rvx,\rvy,\rvz)=q(\rvx,\rvy)q(\rvz|\rvx)=q(\rvx)q(\rvy)q(\rvz|\rvx)$, this is true in case the ground truth factors are the factor of variations (e.g. positions of an object) so that one could vary the factors while keeping the same object. However, for a multi-class labels (e.g. the digits in MNIST), the following assumption is more appropriate: $q(\rvx,\rvy,\rvz)=q(\rvx)q(\rvy|\rvx)q(\rvz|\rvx)$. For our proof, we follow the first case, but changing to the second case is just a matter of switching notation without invalidating the proof.
  \item $q(\rvz|\rvx,\rvy) = \frac{q(\rvx,\rvz|\rvy)q(\rvy)}{q(\rvx,\rvy)}$ (Bayesian theorem).
\end{itemize}
Maximization of ELBO is equivalent to the maximization of
\begin{equation}
\begin{aligned}
    &\quad \expectbig{q(\rvx,\rvy)}{\expectbig{ q(\rvz|\rvx,\rvy) }{ \loge \frac{p(\rvx,\rvy,\rvz)}{q(\rvz|\rvx,\rvy)}} - \log p(\rvx,\rvy)} \\
   =&\quad \expectbig{q(\rvx,\rvy,\rvz)}{\loge \frac{p(\rvx,\rvz|\rvy)p(\rvy)q(\rvx,\rvy)}{q(\rvx,\rvz|\rvy)q(\rvy)p(\rvx,\rvy)}} \\
   =&\quad -\expect{q(\rvy)}{\dkl{q(\rvx,\rvz|\rvy)}{p(\rvx,\rvz|\rvy)}} - \dkl{q(\rvy)}{p(\rvy)} + \dkl{q(\rvx,\rvy)}{p(\rvx,\rvy)}
\end{aligned}
\label{app:eq:maxinfo_sup}
\end{equation}
The result is that $\dkl{q(\rvx,\rvz|\rvy)}{p(\rvx,\rvz|\rvy)} = 0$ for every possible value of $y$, in other words,
$q(\rvx,\rvz) \equiv p(\rvx,\rvz) ~\refstepcounter{equation}(\theequation)\label{app:eq:maxinfo_sup_xz}$.

\textbf{Next we derive the lower bound of $\infop{x}{z}$}
\begin{equation}
\begin{aligned}
       & \dkl{p(\rvx,\rvz)}{q(\rvx,\rvz)} \ge 0 \\
  \iff & \expect{p(\rvx,\rvz)}{\loge p(\rvx,\rvz)} - \expect{p(\rvx,\rvz)}{q(\rvx,\rvz)} \ge 0 \\
  \iff & \underbrace{\expect{p(\rvx,\rvz)}{\loge p(\rvx,\rvz) - \loge p(\rvx) - \loge p(\rvz)}}_{=\infop{x}{z}} \ge
         \expect{p(\rvx,\rvz)}{q(\rvx,\rvz)  - \loge p(\rvx) - \loge p(\rvz)},
\end{aligned}
\label{app:eq:lower_info_xz}
\end{equation}
the bound is exact if $q(\rvx,\rvz) \equiv p(\rvx,\rvz)$.

\textbf{And the lower bound of $\infop{z}{y}$}, using similar approach
\begin{equation}
\begin{aligned}
  \infop{z}{y} \ge \expect{p(\rvz,\rvy)}{q(\rvz,\rvy)  - \loge p(\rvz) - \loge p(\rvy)},
\end{aligned}
\label{app:eq:lower_info_yz}
\end{equation}
the bound is exact if $q(\rvz,\rvy) \equiv p(\rvz,\rvy)$.

The condition that $\infop{x}{z}$ lower bound is maximized is the same as~\eqref{app:eq:maxinfo_sup_xz} (the supervised ELBO), and $\infop{z}{y}$ lower bound is maximized by the same condition in~\eqref{app:eq:maxinfo_uns_yz} (the unsupervised ELBO). According to Theorem~\ref{the:trans_info}, $\infop{x}{y} \ge \infop{x}{z} + \infop{z}{y} - \entropyp{y}$, we conclude that \textit{under optimal conditions (i.e. unlimited amount of data and optimal optimization) SemafoVAE maximizing the lower bound of $\infop{x}{y}$ which encourages the generator to generate more relevant examples associated with the ground truth factors}.

\newpage
\subsection{Training algorithm for SemafoVAE}
\label{app:subsec:semafo_training}

We observe that better predictive model for $q(\rvy|\rvz)$ results small improvement for SemafoVAE, thus, we introduce the supervised loss $\alpha \cdot \loge q(\rvy|\rvz)$ to the supervised ELBO in~\eqref{app:eq:semafo_sup_final}, where $\alpha$ is the scale coefficient fixed to $10$ for all datasets. Similar observation is also mentioned in \cite{kingma14_Semisupervisedlearningdeep}, however, our approach involves oversampling of the labeled data so that the ratio between unsupervised and supervised data within every minibatch is fixed to $10:1$, which explains our choice of $\alpha$ value. Furthermore, we pretrain the reconstruction VAE for 800 iterations without the controller VAE (Figure~\ref{fig:graphical_model}-d) so that $q(\rvy|\rvz)$ gives more stable estimation.

Second, because we repeat the labeled data for oversampling, the algorithm is susceptible to overfitting on supervised examples. This is mitigated by adding extra weight to the reconstruction of unsupervised data, this scale coefficient is $\gamma$ set to $10$ which is the chosen ratio between two data partitions.

There are two sets of parameters for optimization
\begin{itemize}
  \item \textit{Reconstruction VAE}: $\qphi(\rvz|\rvx)$ - encoder, $\pthe(\rvx|\rvz)$ - decoder, and $\qtau(\rvy|\rvz)$ - the predictive factor model
  \item \textit{Controller VAE}:  $\qvf(\rvu|\rvy)$ - encoder, $\ppsi(\rvy|\rvu)$ - decoder, and $\pkap(\rvz|\rvu)$ - the controllable prior
\end{itemize}

\begin{algorithm}[ht]
\caption{SemafoVAE optimization procedure for batch size $m$, and the hyper-parameters: $\alpha$, $\gamma$}
\label{alg:semafovae}
\begin{algorithmic}
  \STATE {\bfseries Input:}
  labeled observation $(x_s^{(i)}, y_s^{(i)})_{i=1}^{\mathcal{S}}$,
  unlabeled observation $(x_u^{(i)})_{i=1}^{\mathcal{U}}$
  \STATE Initialize the networks' parameters: $\{\phi,\theta,\tau\}$ and $\{\varphi,\psi,\kappa\}$
  \REPEAT
    \STATE Random sample a minibatch of size $m$ from $\mathcal{S}$: $(\rvx_s^m, \rvy_s^m)$
    \STATE $\mathbf{g} \gets \nabla_{\phi,\theta,\tau}\big(
      -\elbos(\rvx_s^m,\rvy_s^m,\rvz_s^m;\phi,\theta,\tau) - \alpha\:\loge \qtau(\rvy_s^m|\rvz_s^m) \big)$
      (gradients of minibatch estimator)
    \STATE $\phi,\theta,\tau \gets $ Update parameters using the estimated gradients $\mathbf{g}$
  \UNTIL{number of pretrain steps reached}
  \REPEAT
  \STATE Random sample minibatch of size $m_s=\nicefrac{m}{10}$ from $\mathcal{S}$: $(\rvx_s^{m_s}, \rvy_s^{m_s})$
  \STATE Calculate the ELBO $\elbo(\rvy_s^{m_s},\rvu_s^{m_s};\varphi,\psi)$ and the distribution $\pkap(\rvz_s^{m_s}|\rvu_s^{m_s})$.
  \STATE Calculate the KL-divergence $\dkl{\qphi(\rvz_s^{m_s}|\rvx_s^{m_s})}{\pkap(\rvz_s^{m_s}|\rvu_s^{m_s})}$.
  \STATE $\mathbf{g}_s \gets \nabla_{\phi,\theta,\tau,\varphi,\psi,\kappa}\big(
    -\elbos(\rvx_s^{m_s},\rvy_s^{m_s},\rvz_s^{m_s},\rvu_s^{m_s};\phi,\theta,\tau,\varphi,\psi,\kappa)
    -\alpha\:\loge \qtau(\rvy_s^{m_s}|\rvz_s^{m_s}) \big)$
  \STATE $\phi,\theta,\tau,\varphi,\psi \gets $ Update parameters using the estimated gradients $\mathbf{g}_s$
  \STATE
  \STATE Random sample minibatch of size $m_u=\nicefrac{9m}{10}$ from $\mathcal{U}$: $(\rvx_u^{m_u})$
  \STATE Sampling $\rvy_u^{m_u}$ from the distribution $\qphi(\rvy|\rvz_u^{m_u})$
  \STATE Calculate the ELBO $\elbo(\rvy_u^{m_u},\rvu_u^{m_u};\varphi,\psi)$ and the distribution $\pkap(\rvz_u^{m_u}|\rvu_u^{m_u})$.
  \STATE Calculate the KL-divergence $\dkl{\qphi(\rvz_u^{m_u}|\rvx_u^{m_u})}{\pkap(\rvz_u^{m_u}|\rvu_u^{m_u})}$.
  \STATE $\mathbf{g}_u \gets \nabla_{\phi,\theta,\tau,\varphi,\psi,\kappa}\big(
    -\gamma \; \elbou(\rvx_u^{m_u},\rvz_u^{m_u},\rvu_u^{m_u};\phi,\theta,\tau,\varphi,\psi,\kappa) \big)$
  \STATE $\phi,\theta,\tau,\varphi,\psi \gets $ Update parameters using the estimated gradients $\mathbf{g}_u$
  \UNTIL{$convergence$ of all parameters $\{\phi,\theta,\tau,\varphi,\psi,\kappa\}$}
\end{algorithmic}
\end{algorithm}

\subsection{Implementation details}
\label{app:implementation}

The networks' architecture in Table~\ref{app:tab:networks} are used for all the baselines and our proposed approaches, the architecture is similar to \cite{locatello19_ChallengingCommonAssumptions} and \cite{kingma16_Improvedvariationalinference}. For all datasets, we use Bernoulli distribution to parameterize each pixel independently $\pthe(\rvx|\rvz)=Bernoulli(\rvx|\pi_\theta(\rvz))$. The latent variables are $\mathcal{N}(z|\mu_{\phi}(x),\mathrm{diag}(\sigma_{\phi}(x)))$. For Shapes3D dataset, all factors are discretized, we use Gumbel-Softmax \cite{jang16_CategoricalReparameterizationGumbelSoftmax} for parameterizing every individual factor $y_i$ from the set $\mathcal{Y}$, i.e. $\qtau(y_i|\rvz) = \mathrm{Cat}(y_i|\pi_{\tau}(z))$. For MNIST and FashionMNIST, the one-hot labels are used as factors, and one-hot categorical distribution is used for parameterizing $\rvy$. For MNIST and FashionMNIST, all VAEs have 32 latent units, i.e. $\mathrm{d}_z=32$, this number is chosen based on Figure~\ref{fig:adversarial}-c so that the vanilla VAE with the given architectures is able to converge to its maximum number of active units. For Shapes3D, the CCVAE~\cite{joy21_Capturinglabelcharacteristics} require at least $57$ units (i.e. one unit per discrete value of the factor) for the labeled latents, with an addition of $10$ units for learning the latent styles, in total 67 units are needed. As a result, we use 67 latent units for the whole system, and we also provide results with 10 latent units on Shapes3D in the next Section.

The FactorVAE discriminator and its hyperparameters are the same as described in \cite{kim18_Disentanglingfactorising}. For hierarchical VAE, we use bidirectional inference as in \cite{kingma16_Improvedvariationalinference}, only one extra latent layer is added which consists of 64 units for MNIST and 128 units for Shapes3D. For SemafoVAE, a linear fully connected network is used to project $\rvz$ to $\rvy$ in $\qtau(\rvy|\rvz)$ so to ensure maximum association between $\rvy$ and $\rvz$. A similar approach applied to $\pkap(\rvz|\rvu)$ (the controllable prior). The architecture of the \textit{ControllerVAE} are in Table~\ref{app:tab:controller_vae} which is chosen without any fine-tuning. We select $\beta=10$ for BetaVAE and $\gamma=10$ for GammaVAE.

All networks are trained using Adam optimizer \cite{kingma17_AdamMethodStochastic} with learning rate $10^{-3}$ for MNIST, FashionMNIST and  $10^{-4}$ for Shapes3D. We set batch size to 64, and the maximum iteration for each training to \textit{200,000} iterations for MNIST, FashionMNIST and \textit{2,000,000} for Shapes3D. This number is guaranteed for all systems to converge to their best performance, and during training, only best-performed weights (on validation set) are saved.

\textbf{Computational resources} Our resources are limited, most experiments were run on GTX 1080 GPU. Training consumed $\sim 1Gb$ of GPU memory for MNIST and FashionMNIST and $\sim 4Gb$ for Shapes3D. For \textit{200,000} iterations on MNIST, the algorithm took $\sim 3$ hours. For Shapes3D, it took $\sim 16$ hours to run two million iterations. The difference in training time among algorithms is trivial\footnote{All the code and running configurations will be available on Github provided under the MIT license}.

\begin{table}[ht]
\vspace{-10pt}
\caption{Encoder and Decoder architecture for MNIST, FashionMNIST (left) and Shapes3D (right), where \textit{dense} indicates fully connected feedforward neural network, $a \times b \times c$ \textit{conv} is a convolutional neural network with $(a, b)$ kernel size and $c$ filters, \textit{deconv} is the transposed convolutional neural network, $\mathrm{d}_z$ is the number of latent units. We use exponential linear unit (ELU) instead of ReLU as recommended in \cite{kingma16_Improvedvariationalinference}.}
\label{app:tab:networks}
\begin{center}
\begin{adjustbox}{max width=\textwidth}
\begin{tabular}[c]{llll}
\toprule
\multicolumn{2}{c}{\textbf{MNIST} \& \textbf{FashionMNIST}} & \multicolumn{2}{c}{\textbf{Shapes3D}} \\
\textbf{Encoder} & \textbf{Decoder}                & \textbf{Encoder} & \textbf{Decoder}\\
\midrule
Input $28 \times 28$ gray image          & Input $\in \mathbb{R}^{\mathrm{d}_z}$                                & Input $64 \times 64$ RGB image            & Input $\in \mathbb{R}^{\mathrm{d}_z}$ \\
Normalize pixels $[-1, 1]$                            & \textit{dense} 196, Linear, reshape $(7, 7, 4)$         & Normalize pixels $[-1, 1]$             & \textit{dense} 256, Linear, reshape $(4, 4, 16)$ \\
$5 \times 5 \times 32$ \textit{conv}, ELU, stride $1$ & $5 \times 5 \times 64$ \textit{deconv}, ELU, stride $2$ & $4 \times 4 \times 32$ \textit{conv}, ELU, stride $2$ & $4 \times 4 \times 64$ \textit{deconv}, ELU, stride $2$ \\
$5 \times 5 \times 32$ \textit{conv}, ELU, stride $2$ & $5 \times 5 \times 64$ \textit{deconv}, ELU, stride $1$ & $4 \times 4 \times 32$ \textit{conv}, ELU, stride $2$ & $4 \times 4 \times 64$ \textit{deconv}, ELU, stride $2$ \\
$5 \times 5 \times 64$ \textit{conv}, ELU, stride $1$ & $5 \times 5 \times 32$ \textit{deconv}, ELU, stride $2$ & $4 \times 4 \times 64$ \textit{conv}, ELU, stride $2$ & $4 \times 4 \times 32$ \textit{deconv}, ELU, stride $2$ \\
$5 \times 5 \times 64$ \textit{conv}, ELU, stride $2$ & $5 \times 5 \times 32$ \textit{deconv}, ELU, stride $1$ & $4 \times 4 \times 64$ \textit{conv}, ELU, stride $2$ & $4 \times 4 \times 32$ \textit{deconv}, ELU, stride $2$ \\
\textit{dense} 196, Linear                            & $1 \times 1 \times 1$ \textit{conv}, Linear, stride $1$ & \textit{dense} 256, Linear & $1 \times 1 \times 3$ \textit{conv}, Linear, stride $1$ \\
                                                      & Bernoulli(logits=$x$)                                   &                            & Bernoulli(logits=$x$) \\
\bottomrule
\end{tabular}
\end{adjustbox}
\end{center}
\vspace{-10pt}
\end{table}

\begin{table}[ht]
\vspace{-10pt}
\caption{Controller VAE for SemafoVAE, where $\mathrm{d}_y$ is the total number of dimension for ground truth factors (10 for MNIST and FashionMNIST; 57 for Shapes3D).}
\label{app:tab:controller_vae}
\begin{center}
\begin{adjustbox}{max width=0.6\textwidth}
\begin{tabular}[c]{lcl}
  \toprule
  \textbf{Encoder} && \textbf{Decoder} \\
  \midrule
  Input $\in \mathbb{R}^{\mathrm{d}_y}$ && Input $\in \mathbb{R}^{\mathrm{d}_y}$\\
  dense 512, ReLU && dense 512, ReLU \\
  dense 512, ReLU && dense 512, ReLU \\
                  && dense $\mathrm{d}_y$, Linear \\
                  && $\prod_{y_i\in\mathcal{Y}}\mathrm{GumbelSoftmax}(y_i|\mathrm{logits}=x_i)$ \\
  \bottomrule
\end{tabular}
\end{adjustbox}
\end{center}
\vspace{-10pt}
\end{table}

\subsection{Additional Experiments and Results}
\label{app:additional}

\textbf{Note on calculating the Fr{\'e}chet Inception Distance for semi-supervised VAE}. For a model with controllable generation, i.e. CCVAE (\cite{joy21_Capturinglabelcharacteristics}) and SemafoVAE, generate complete random samples is an issue since the model needs to know which factors to be generated. Our approach in Table~\ref{tab:llk} is that \textit{repeating the same set of sampled factors in every minibatch for generation}, however, the FID as a measure of distance between two distributions gives a lower score to this approach. We suspect that the generated examples need to cover the whole distribution and the model must ensure the diversity of the generated samples. As a result, we \textit{randomize the new set of factors for every minibatch when generating examples for FID}, the FID for CCVAE improves from \textbf{115.17} to \textbf{83.72}, and the FID for SemafoVAE improves from \textbf{92.28} to \textbf{28.62} on Shapes3D dataset which is the best FID among all models. The FID scores in this section are reported based on the second method.

\subsubsection{Varying the supervision rate for SemafoVAE}

\begin{figure}[ht]
\begin{small}
  \begin{center}
    \includegraphics[width=\linewidth]{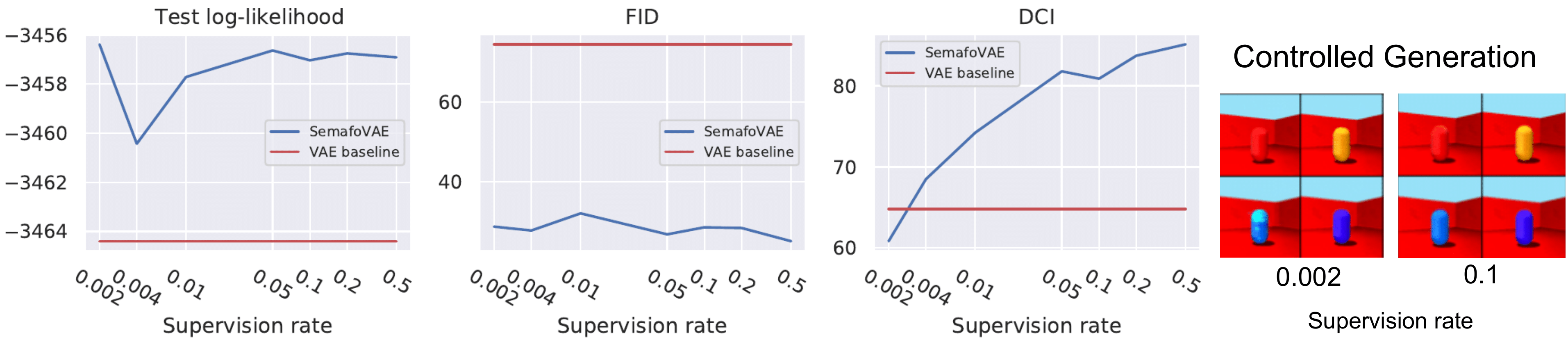}
  \end{center}
  \caption{Performance of SemafoVAE with different supervision rate on Shapes3D dataset.}
  \label{app:fig:supervision_rate}
\end{small}
\end{figure}

While no significant improvement is achieved for a greater than $0.1$ supervision rate, the performance of SemafoVAE is consistent among all configurations Figure~\ref{app:fig:supervision_rate}. As small as $0.004$ percent of supervision data is enough to improve the general performance and gain control of the generation, however, artifacts are observed in the controlled generation of the model with lower supervision rate, e.g. $0.002$ (Figure~\ref{app:fig:supervision_rate}).

\subsubsection{Comparing the sampled images from the latents' prior distribution}

\begin{figure}[ht]
\begin{small}
  \begin{center}
    \includegraphics[width=\linewidth]{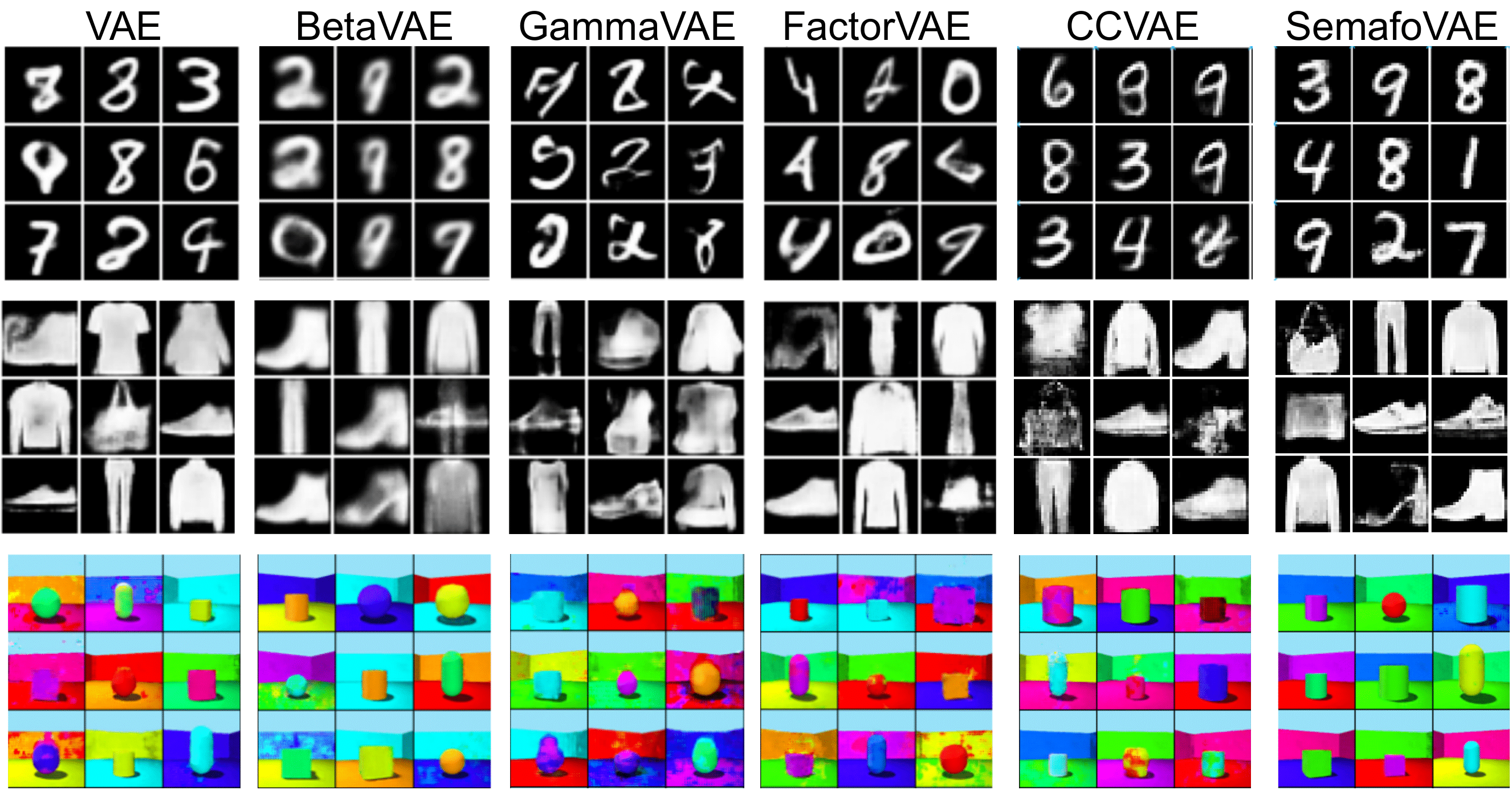}
  \end{center}
  \caption{Randomly sampled images using the prior distribution of the latent units. \textit{SemafoVAE generates shaper images than any existing methods}.}
  \label{app:fig:random_samples_prior}
\end{small}
\end{figure}

For the unsupervised methods, we draw samples directly from the latent prior distributions, then using the decoder to reconstruct the output images. For the semi-supervised methods (CCVAE and SemafoVAE), first, we sample the factors (the class labels for MNIST and Fashion MNIST; the factor of variation for Shapes3D), then we acquire the latent prior distributions given the factors, and finally, we draw samples from the prior distributions and reconstruct the images using the decoder. Results are showed in Figure~\ref{app:fig:random_samples_prior}.

\subsubsection{Comparing the traverse of the latent posterior distribution}

\begin{figure}[!htbp]
\vspace{-5pt}
\begin{small}
\begin{center}
  \includegraphics[width=\linewidth]{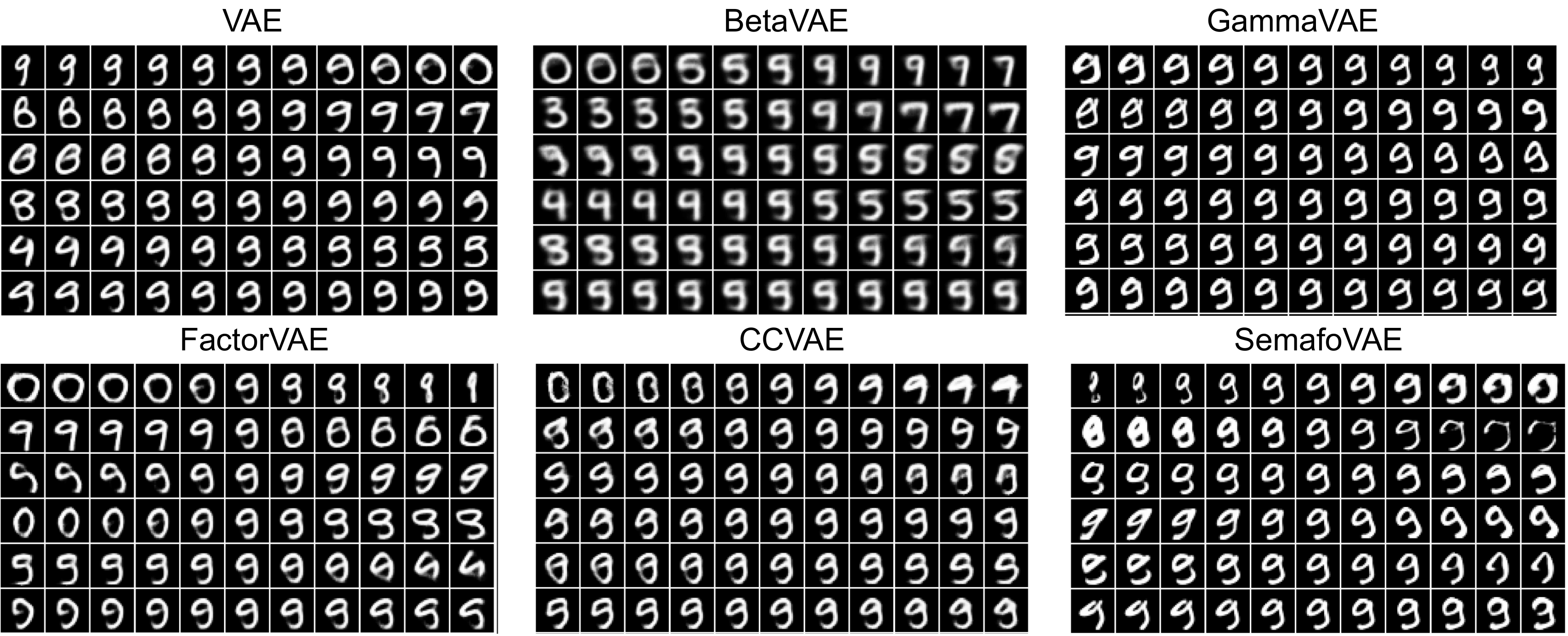}
\end{center}
\caption{Posterior traverse on MNIST dataset. Both CCVAE and SemafoVAE were able to capture the style of number ``9'', however, SemafoVAE learns more factor of variation of the given number.}
\label{app:fig:post_mnist}
\end{small}
\vspace{-5pt}
\end{figure}

\begin{figure}[!htbp]
\vspace{-5pt}
\begin{small}
\begin{center}
  \includegraphics[width=\linewidth]{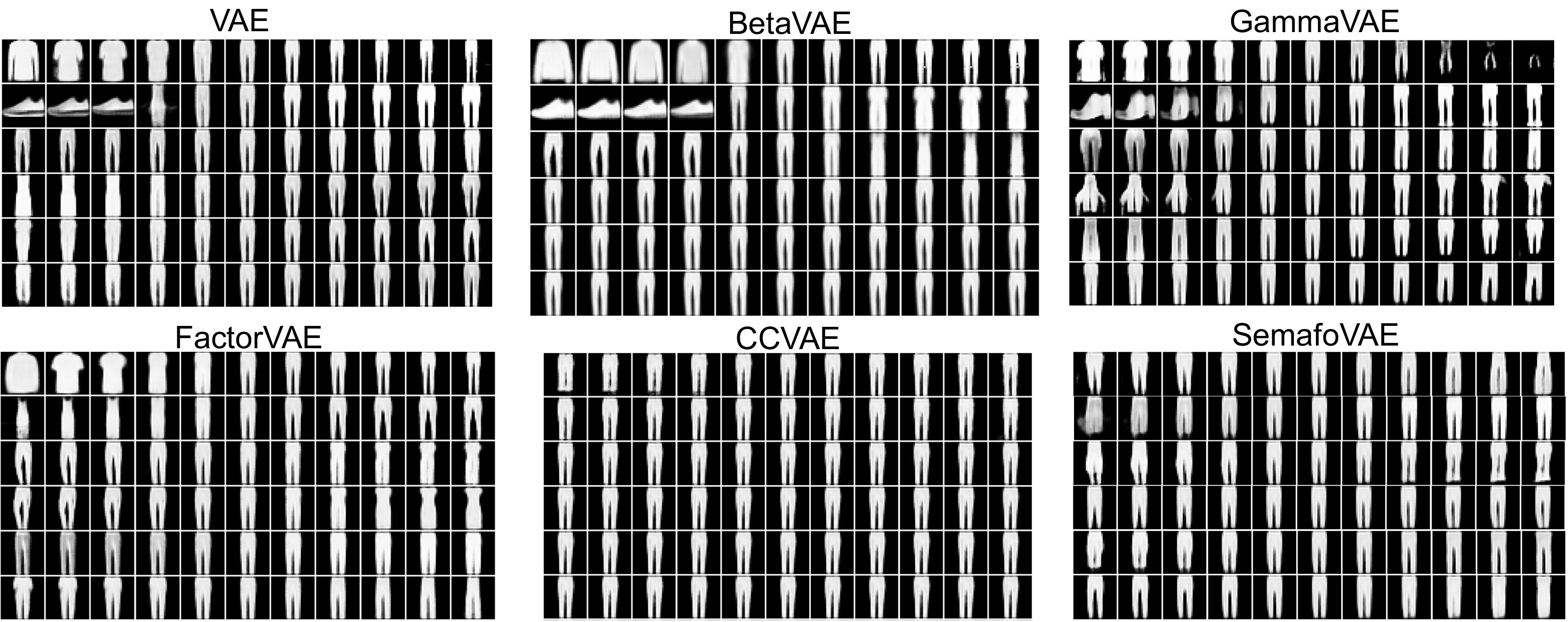}
\end{center}
\caption{Posterior traverse on Fashion MNIST dataset. All unsupervised methods cannot disentangle the type of clothes from the style of clothes. SemafoVAE can capture distinguished features of a ``trouser'' (the first row interestingly shows the traverse from a sport tight to a disco jean).}
\label{app:fig:post_fmnist}
\end{small}
\vspace{-5pt}
\end{figure}

\begin{figure}[!htbp]
\vspace{-5pt}
\begin{small}
\begin{center}
  \includegraphics[width=\linewidth]{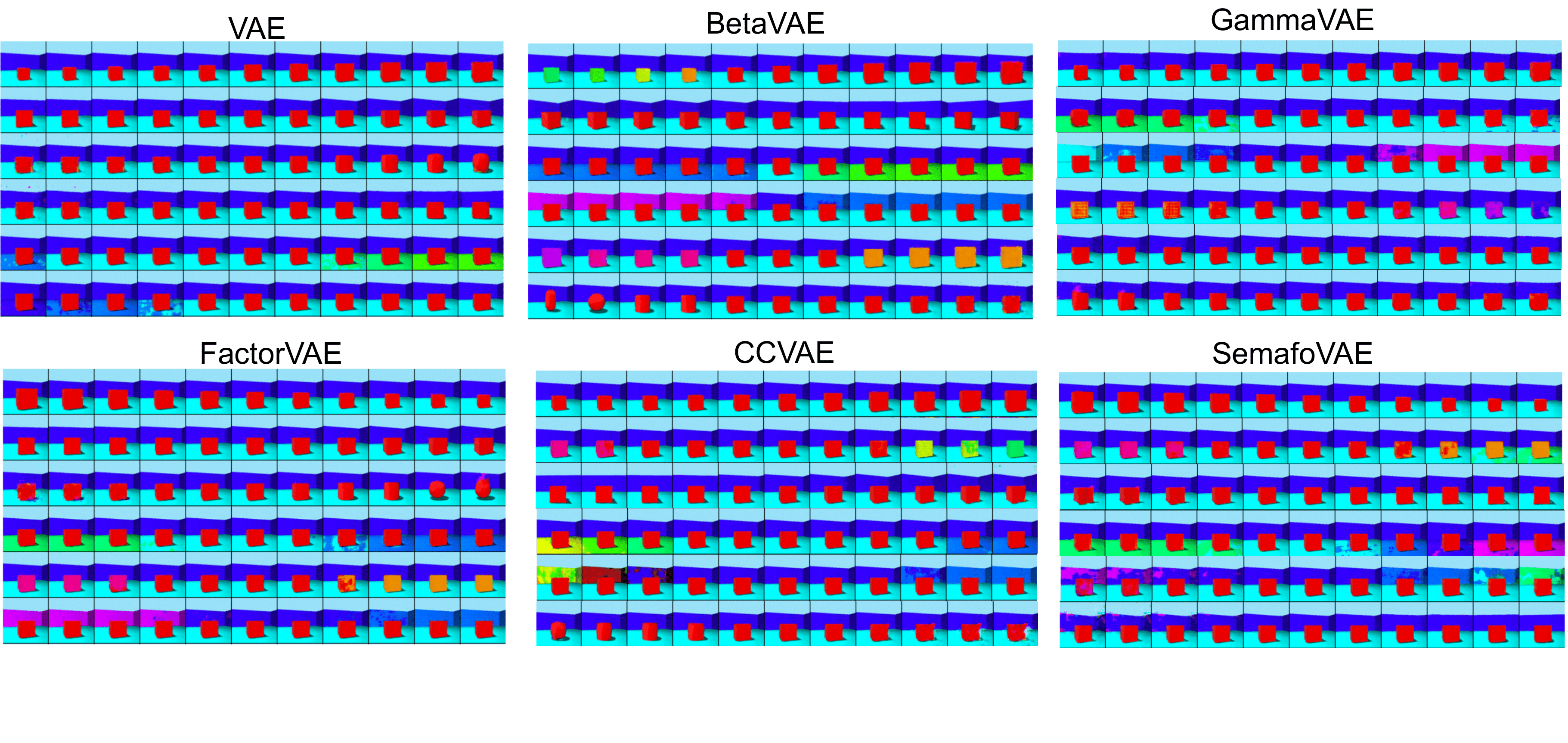}
\end{center}
\caption{Posterior traverse on Shapes3D dataset. All unsupervised methods show a certain level of disentanglement which encapsulate more than one factor into a single latent dimension. Both CCVAE and SemafoVAE can disentangle the factors, i.e. the shape remains invariant during the traverse while scale, orientation, wall hue, floor hue, and object hue are changing. Notably, only the semi-supervised methods capture the room orientation in the traverse (third row for CCVAE and sixth row for SemafoVAE).}
\label{app:fig:post_shapes3d}
\end{small}
\vspace{-5pt}
\end{figure}

The \textit{posterior traverse} experiment is performed by selecting a random example from the test set, then extracting its latent representation. Next, we linearly traverse each latent dimension from $-2.5\sigma$ to $2.5\sigma$ around its mean value and using the decoder for reconstructing the images. For all the posterior traverse figures, we select the top 6 most variate latent dimensions that correlated to the ground truth factors. Results are showed in Figure~\ref{app:fig:post_mnist}, Figure~\ref{app:fig:post_fmnist} and Figure~\ref{app:fig:post_shapes3d}.

\newpage
\subsubsection{Comparing the traverse of latents' prior distribution}

\begin{figure}[!htbp]
\vspace{-5pt}
\begin{small}
\begin{center}
  \includegraphics[width=\linewidth]{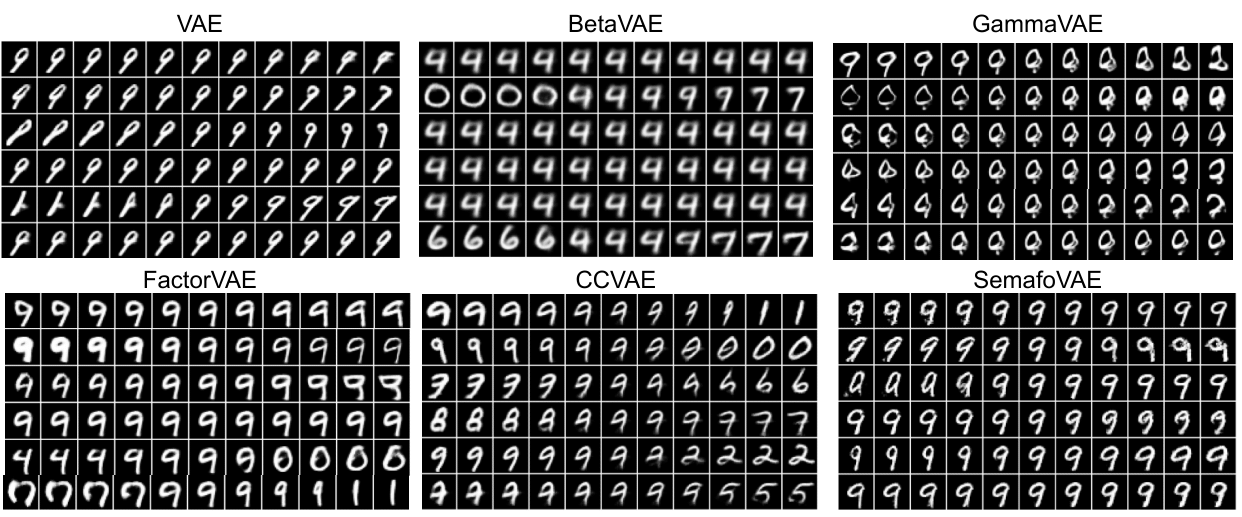}
\end{center}
\caption{Traverse on the latents' prior distribution for MNIST dataset. All unsupervised methods show a mixing of styles and digits during the traverse of all dimensions. CCVAE shows certain variations in digit style, however, the images are transformed into different numbers at the end of the spectrum. SemafoVAE shows consistently in the presented number while traversing the styles of number ``9''.}
\label{app:fig:prior_mnist}
\end{small}
\vspace{-5pt}
\end{figure}

\begin{figure}[!htbp]
  \vspace{-5pt}
  \begin{small}
\begin{center}
  \includegraphics[width=\linewidth]{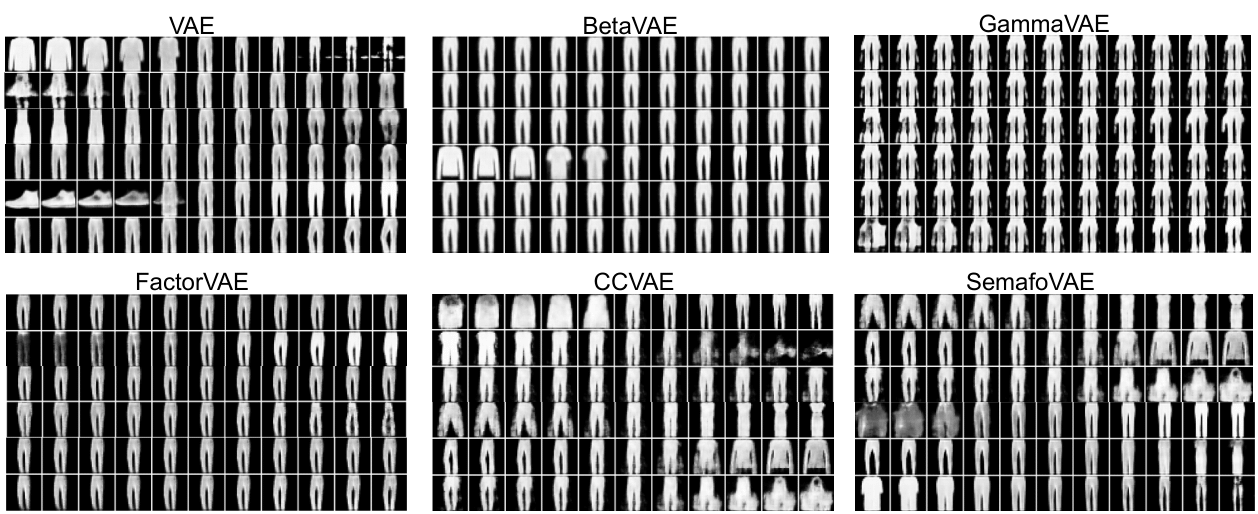}
\end{center}
\caption{Traverse on the latents' prior distribution for Fashion MNIST dataset. FactorVAE is the only unsupervised method that captured style variations of the trouser without confusing the label. Both CCVAE and SemafoVAE show class confusion when traversing too far from the mean values.}
\label{app:fig:prior_fmnist}
\end{small}
\vspace{-5pt}
\end{figure}

\begin{figure}[ht]
\begin{small}
\begin{center}
  \includegraphics[width=\linewidth]{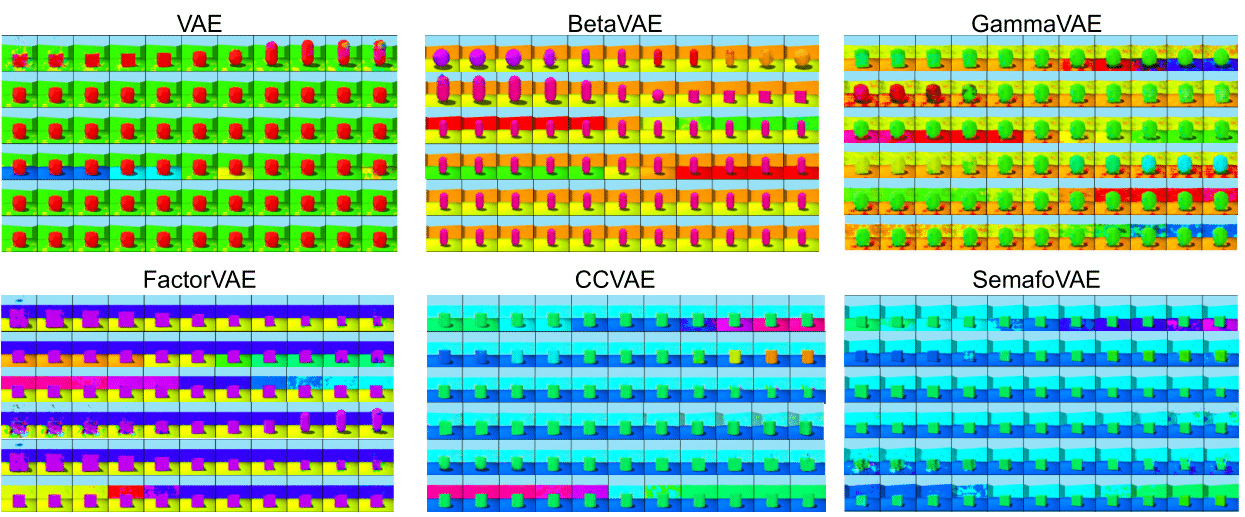}
\end{center}
\caption{Traverse on the latents' prior distribution for Shapes3D dataset. SemafoVAE shows the best quality images, and is the only method able to capture object-orientation (the fifth row) separately from the room orientation (the fourth row).}
\label{app:fig:prior_shapes3d}
\end{small}
\end{figure}

For this experiment, a random sample is drawn from the prior distribution. Then we use this vector as a reference and applying linear traversal for each of it dimension which results in a series of new latent representation. Finally, we use the decoder to reconstruct the image from the traverse vectors. For MNIST and Fashion MNIST, since we cannot know in advance which class will be generated using the unsupervised method, we perform sampling until we saw the class of interest (``9'' for MNIST and ``trouser'' for Fashion MNIST). For Shapes3D, no particular filtering was performed, as a result, the images show different objects with different factors of variation. Results are showed in Figure~\ref{app:fig:prior_mnist}, Figure~\ref{app:fig:prior_fmnist} and Figure~\ref{app:fig:prior_shapes3d}.

Since CCVAE has two latent spaces for style and class, we only perform prior traverse on the prior distributions of style latents.

\end{document}